\documentclass[letterpaper]{article} 
\usepackage{aaai2027}  
\usepackage[hyphens]{url}  
\usepackage{graphicx} 
\usepackage{natbib}  
\usepackage{caption} 
\usepackage{algorithm}
\usepackage{algpseudocode}
\usepackage{multirow}
\usepackage{amsmath}

\usepackage{newfloat}
\usepackage{listings}
\usepackage{subcaption}
\DeclareCaptionStyle{ruled}{labelfont=normalfont,labelsep=colon,strut=off} 
\floatstyle{ruled}
\newfloat{listing}{tb}{lst}{}
\floatname{listing}{Listing}

\usepackage{booktabs}
\usepackage{amssymb}
\title{ParetoPilot: Zero-Surrogate Offline Multi-Objective Optimization via Infer-Perturb-Guide Diffusion}
\author{
    Ruiqing Sun\textsuperscript{\rm 1},
    Sen Yang\textsuperscript{\rm 2},
    Dawei Feng\textsuperscript{\rm 1},
    Bo Ding\textsuperscript{\rm 1},
    Yijie Wang\textsuperscript{\rm 1}\thanks{Corresponding author.},
    Huaimin Wang\textsuperscript{\rm 1}
}
\affiliations{
    \textsuperscript{\rm 1}College of Computer Science and Technology, National University of Defense Technology, Changsha 410073, China\\
    \textsuperscript{\rm 2}Bioinformatics Center of AMMS, Beijing 100080, China\\
    
    sunny0331@foxmail.com, yangsen.nudt@hotmail.com, davyfeng.c@qq.com, \{dingbo, wangyijie, hmwang\}@nudt.edu.cn
}

\begin{document}

\maketitle

\begin{abstract}
Offline multi-objective optimization (Offline MOO) seeks Pareto-optimal designs from static datasets without additional environment interactions. Existing generative methods typically guide sampling with external surrogate or preference models, which adds training complexity and may provide unreliable guidance. 
We propose ParetoPilot, a plug-and-play method that guides designs to Pareto front at inference time using a pre-trained conditional diffusion model without any surrogate. 
ParetoPilot introduces an Infer-Perturb-Guide (IPG) engine within the reverse diffusion process. IPG first infers the individual conditional target for each sample in the batch by aligning its conditional and unconditional predictions. It then perturbs these targets collectively across the batch, balancing convergence toward the Pareto front and diversity among samples. Finally, the engine guides the generative trajectory toward the Pareto front by injecting these perturbed targets via standard Classifier-Free Guidance (CFG).
Experiments on 51 tasks demonstrate that ParetoPilot achieves the best overall ranking among 16 methods and competitive hypervolume improvement.

\end{abstract}


\section{Introduction}
\label{sec:intro}
Offline MOO is ubiquitous in real-world problems. Rather than simply screening existing samples, offline MOO learns the underlying joint distribution from a static, pre-collected dataset. It then generates novel Pareto-optimal designs through extrapolation or interpolation without further environment interactions. This paradigm is essential in real-world applications such as drug discovery \cite{antoniuk2026active}, material design \cite{fromer2023computer}, and hardware architecture search \cite{lu2023neural}, where online function evaluations are prohibitively expensive or strictly limited. Unlike single-objective optimization, MOO requires simultaneously satisfying two inherently conflicting goals: Convergence (pushing solutions toward the unobserved true Pareto Front) and Diversity (ensuring uniform and broad coverage across the objective trade-off space).

To tackle these challenges, recent advances have predominantly adopted an external surrogate paradigm. Inverse generative approaches based on Flow Matching or Diffusion Models (e.g., ParetoFlow \cite{paretoflow}, PGD-MOO \cite{pgd}) achieve state-of-the-art results by relying on external proxy regressors or preference classifiers to guide generation. However, this reliance on auxiliary models introduces three bottlenecks. First, training a separate surrogate per objective incurs substantial computational overhead and extensive hyperparameter tuning. Second, these proxies are unreliable during extrapolation, especially at the boundaries of the data range, where they output deceptive scores for invalid out-of-distribution designs. Third, training proxies requires continuous access to the original offline dataset. This requirement conflicts with the contemporary paradigm of deploying models solely as weights \cite{bommasani2021opportunities} and is often infeasible under strict data privacy regulations \cite{kaissis2020secure}. Although the recent surrogate-free method \cite{shrestha2026pareto} bypasses external proxies, it still requires raw dataset access for Pareto-oriented reweighting. This training scheme may degrades generation quality and lacks flexibility, requiring full retraining whenever optimization preferences change. Furthermore, its static targets are heuristically inferred from offline data rather than the learned  Pareto front, often yielding conservative or physically implausible designs.

To overcome these bottlenecks, we propose \textbf{ParetoPilot}. It requires no external proxy model and avoids raw dataset access by operating solely on a pre-trained conditional diffusion model. The core of ParetoPilot is an Infer-Perturb-Guide (IPG) engine, which progressively steers samples toward the Pareto front by updating the conditioning targets and CFG strength during sampling.
To preserve design plausibility, IPG is interleaved with unconditional denoising, allowing the diffusion model to keep samples close to the learned data manifold. At each IPG step, it first \textbf{infers} a conditioning direction aligned with each sample's current unconditional denoising trajectory. It then \textbf{perturbs} the inferred condition across the batch by balancing convergence toward the Pareto front with repulsion among samples for diversity. To dynamically balance these two goals throughout denoising, the convergence and diversity components are orthogonalized and weighted by a time-dependent schedule, yielding optimized conditioning targets. Finally, these targets are injected back through CFG to \textbf{guide} the samples. Since ParetoPilot only modifies the CFG targets and strengths without changing the sampling procedure, it serves as a plug-and-play inference-time method for conditional diffusion models.

Our main contributions are summarized as follows:
\begin{itemize}
    \item We propose \textbf{ParetoPilot}, a plug-and-play inference-time method that turns a pre-trained conditional diffusion model into an optimizer for offline MOO. By steering generation through the existing CFG interface, ParetoPilot discovers Pareto-optimal designs without auxiliary proxy models, retraining, or deployment-time access to the original offline dataset.

    \item We introduce the IPG engine, which dynamically reshapes the generative trajectory using only the conditional priors internal to the diffusion model. By inferring, perturbing, and re-injecting the denoising directions, it steers sampling toward designs that are simultaneously well-converged and well-distributed across promising Pareto regions.
    \item Extensive experiments across 51 synthetic and real-world tasks against 15 baselines demonstrate that ParetoPilot achieves strong overall performance with robust Pareto-front coverage.
\end{itemize}

\section{Background}
\label{sec:background}

\subsection{Offline Multi-Objective Optimization}
Without loss of generality, we consider a minimization problem involving $m$ distinct and potentially conflicting objective functions. Let $\mathcal{X} \subset \mathbb{R}^d$ denote the design space and $\mathcal{Y} \subset \mathbb{R}^m$ denote the objective space. The goal of MOO is to optimize a vector-valued black-box function $f(x) = [f_1(x), \dots, f_m(x)]^\top$. MOO relies on the concept of Pareto dominance to compare different designs.

A design $x_A$ is said to Pareto dominate another design $x_B$ (denoted as $x_A \prec x_B$) if $f_i(x_A) \le f_i(x_B)$ for all $i \in \{1, \dots, m\}$ and there exists at least one index $j$ such that $f_j(x_A) < f_j(x_B)$. A design $x^*$ is considered Pareto-optimal if no other design in the feasible space can Pareto dominate it. The set of all Pareto-optimal designs is known as the Pareto set, and their corresponding coordinates in the objective space constitute the Pareto front. 

In the offline setting, the true objective function $f(x)$ is entirely inaccessible for online queries. Instead, the optimization algorithm is strictly constrained to a pre-collected, static dataset $\mathcal{D} = \{(x^{(i)}, y^{(i)})\}_{i=1}^N$, where $x^{(i)} \in \mathcal{X}$ is a previously evaluated design and $y^{(i)} = f(x^{(i)}) \in \mathcal{Y}$ is its corresponding objective vector. Although the empirical Pareto front of the offline dataset, denoted by $y_{\mathcal{D},\mathrm{pareto}}$, is accessible, it is generally not the true Pareto front. A well-trained generative model may interpolate and extrapolate beyond the observed dataset, so directly conditioning on $y_{\mathcal{D},\mathrm{pareto}}$ or on a fixed extreme target can be suboptimal. Therefore, offline MOO should be formulated as a step-by-step process that dynamically navigates the latent condition space toward promising Pareto regions. Consequently, generative methods cannot simply condition the sampling process on $y_{\mathcal{D}, pareto}$ or an arbitrary static extreme target. Instead, the optimization must be formulated as a step-by-step gradual approach to dynamically navigate the latent space and approximate the implicitly learned $y_{pareto}$.

\subsection{Conditional Diffusion Models and Classifier-Free Guidance}
Diffusion models \cite{ddpm} are a class of generative models that learn to synthesize data through a parameterized Markov chain. The forward process gradually corrupts a data sample $x_0 \sim q(x)$ by injecting Gaussian noise over $T$ timesteps, producing a sequence of increasingly noisy latent states $x_1, \dots, x_T$. The reverse denoising process aims to iteratively recover the original data from a standard Gaussian noise distribution $x_T \sim \mathcal{N}(0, I)$ using a neural network $\epsilon_\theta$ trained to predict the injected noise.

To enable conditional generation, the denoising neural network is typically conditioned on an external signal $y$, forming the conditional noise predictor $\epsilon_\theta(x_t, y, t)$. A ubiquitous strategy to enhance the alignment between the generated samples and the given condition is CFG \cite{cfg}. During the training phase of a CFG-enabled diffusion model, the condition $y$ is randomly replaced with a null token $\emptyset$ with a predefined dropout probability, allowing the single network to simultaneously learn both the conditional distribution $\epsilon_\theta(x_t, y, t)$ and the unconditional data distribution $\epsilon_\theta(x_t, \emptyset, t)$. 
During the sampling phase, CFG changes the noise prediction by pushing the conditional estimate away from the unconditional estimate. The modified noise prediction $\tilde{\epsilon}_t$ is computed as follows:
\begin{equation}
    \tilde{\epsilon}_t = \epsilon_\theta(x_t, \emptyset, t) + w \cdot \left( \epsilon_\theta(x_t, y, t) - \epsilon_\theta(x_t, \emptyset, t) \right)
    \label{eq:cfg}
\end{equation}
where $w$ is the guidance scale. 

In standard supervised generation tasks, the target condition $y$ is explicitly provided by the user. However, in offline MOO, the optimal objective coordinate $y_{pareto}$ corresponding to a promising but unobserved design is inherently unknown.

\section{Related Work}
\label{sec:related_work}

\textbf{Online Multi-Objective Optimization.} 
Adaptive experimental design has extensively explored MOO in an online sequential fashion. The most established framework in this domain is Multi-Objective Bayesian Optimization (MOBO), which iteratively updates probabilistic surrogate models using newly acquired environment data to guide the search for the Pareto front. Typical implementations include the standard MOBO-Vanilla based on expected hypervolume improvement \cite{daulton2021parallel}, MOBO-ParEGO utilizing random scalarizations to decompose the objectives \cite{knowles2006parego}, and MOBO-JES employing information-theoretic joint entropy search \cite{hvarfner2022joint}. While highly effective in data-scarce regimes, these sequential approaches require continuous queries to the black-box environment. Consequently, they are entirely inapplicable to offline settings.

\textbf{Offline Multi-Objective Optimization.} 
To address the critical limitations of online queries, offline MOO restricts the optimization process entirely to a static, pre-collected dataset. The conventional forward paradigm involves training deep neural networks as proxy evaluators on the offline data, followed by applying evolutionary search algorithms like NSGA-II \cite{deb2002nsgaii} to find optimal designs. Recent benchmarking efforts \cite{xue2024offline} categorize these forward surrogates into End-to-End architectures, Multi-Head models, and Multiple Independent Models. To mitigate the inherent inaccuracies of proxy models, various advanced regularization and loss balancing techniques have been widely integrated. These include gradient manipulation methods such as GradNorm \cite{chen2018gradnorm} and PcGrad \cite{yu2020gradient} applied to End-to-End and Multi-Head architectures. Furthermore, robust adaptation and self-training strategies, including Conservative Objective Models (COMs) \cite{trabucco2021conservative}, RoMA \cite{yu2021roma}, IOM \cite{qi2022data}, ICT \cite{yuan2023importance}, and Tri-Mentoring \cite{chen2023parallel}, are frequently employed alongside Multiple Models to stabilize the predictions. Despite these extensive enhancements, forward methods suffer significantly from out-of-distribution vulnerability. When the search algorithm extrapolates beyond the training distribution to discover better designs, these static surrogates inevitably produce overconfident and erroneous predictions, leading to invalid solutions and manifold collapse.

\textbf{Generative Inverse Approaches.} 
Inverse generative approaches attempt to model the conditional distribution of high-quality designs directly. ParetoFlow \cite{paretoflow} proposes a guided flow-matching framework, where the generation process is steered by scalarized gradients from separately trained surrogate models combined with uniformly sampled preference weights. PGD-MOO \cite{pgd} adopts a classifier-guided diffusion framework, in which an external preference classifier is trained on pairs of noised offline samples to predict Pareto dominance, with crowding-distance information incorporated to promote diversity. More recently, PCD \cite{shrestha2026pareto} removes external neural surrogates by training a conditional diffusion model with a Pareto-oriented reweighted denoising objective, and performs inference using target conditions constructed from the static offline dataset.

Despite their strong empirical performance, these generative methods still suffer from several structural limitations. ParetoFlow and PGD-MOO rely on external surrogate or preference models, which introduces additional training cost and exposes generation to proxy approximation errors outside the data distribution. PCD avoids external neural proxies, but still requires Pareto-oriented reweighted training of the generative model. This limits flexibility, since the model may need to be retrained when optimization preferences change, and may compromise generation quality by biasing the learned distribution. In addition, its static dataset-derived targets and Gaussian-perturbed diversity conditions are not guaranteed to match the conditional distribution learned by the diffusion model, potentially affecting the plausibility and convergence of generated solutions. Finally, PCD still requires raw dataset access, leaving the deployment challenge unresolved when training data are restricted or unavailable.

\section{Method}
\label{sec:method}

\begin{algorithm}[tb] 
\caption{ParetoPilot: IPG Engine for Offline MOO}
\label{alg:igp}
\small 
\begin{algorithmic}[1]
\Require Diffusion $\epsilon_\theta$, proposals $N$, steps $T$, probes $k$, warmup $T_{warmup}$
\State $x_T \sim \mathcal{N}(\mathbf{0}, \mathbf{I}), \quad y \leftarrow \mathbf{0.5} \in \mathbb{R}^{N \times m}$
\State $\text{Opt} \leftarrow \text{Adam}(y, \text{lr}=\eta)$ \Comment{Global un-reset optimizer}
\For{$t = T, T-1, \dots, 1$}
    \State $\epsilon_{\emptyset} \leftarrow \epsilon_\theta(x_t, \emptyset, t)$ 
    \If{$t \equiv 0 \pmod V$ \textbf{or} $t \in \{1, T-1\}$}
        \For{$1$ \textbf{to} $k$} \Comment{\textbf{Step 1: INFER}}
            \State $\mathcal{L}_{align} \leftarrow \| \epsilon_\theta(x_t, y, t) - \epsilon_{\emptyset} \|_2^2$
            \State $y \leftarrow \text{Opt.step}(\nabla_y \mathcal{L}_{align})$ \Comment{Keep EMA momentum}
        \EndFor
        \State $y^* \leftarrow \text{detach}(y)$
        
        \If{$t \le T - T_{warmup}$} \Comment{\textbf{Step 2: PERTURB}}
            \State $\tau_t \leftarrow t / (T - T_{warmup})$ \Comment{Linearly decays to $0$}
            \State $\alpha_t, w_t \leftarrow \text{LinearSchedule}(\tau_t)$
            \State $\vec{d}_{conv} \leftarrow -\mathbf{1} / \sqrt{m}$ \Comment{Parallel gravity}
            \State Compute repulsion $\vec{F}_{rep}$ and edgeness $\mathcal{E}$ for $y^*$
            \State $\vec{F}_{\perp} \leftarrow \vec{F}_{rep} - (\vec{F}_{rep} \cdot \vec{d}_{conv}) \vec{d}_{conv}$ \Comment{Gram-Schmidt}
            \State $\vec{d}_{div} \leftarrow \vec{F}_{\perp} / \|\vec{F}_{\perp}\|_2$
            \State $\vec{d}_{target} \leftarrow \text{Norm}\left(\alpha_t \vec{d}_{conv} + (1-\alpha_t) \mathcal{E} \vec{d}_{div}\right)$
            \State $y^{target} \leftarrow y^* + \gamma \cdot \vec{d}_{target}$
        \Else \Comment{Warmup Phase}
            \State $y^{target} \leftarrow y^*, \quad w_t \leftarrow 0.0$
        \EndIf
    \EndIf
    \State $\tilde{\epsilon}_t \leftarrow \epsilon_{\emptyset} + w_t \cdot ( \epsilon_\theta(x_t, y^{target}, t) - \epsilon_{\emptyset} )$ \Comment{\textbf{Step 3: GUIDE}}
    \State $x_{t-1} \leftarrow \text{DDPM\_Step}(x_t, \tilde{\epsilon}_t, t)$
\EndFor
\State \Return $x_0$
\end{algorithmic}
\end{algorithm}

\begin{figure*}[t]
    \centering
    \includegraphics[width=\textwidth]{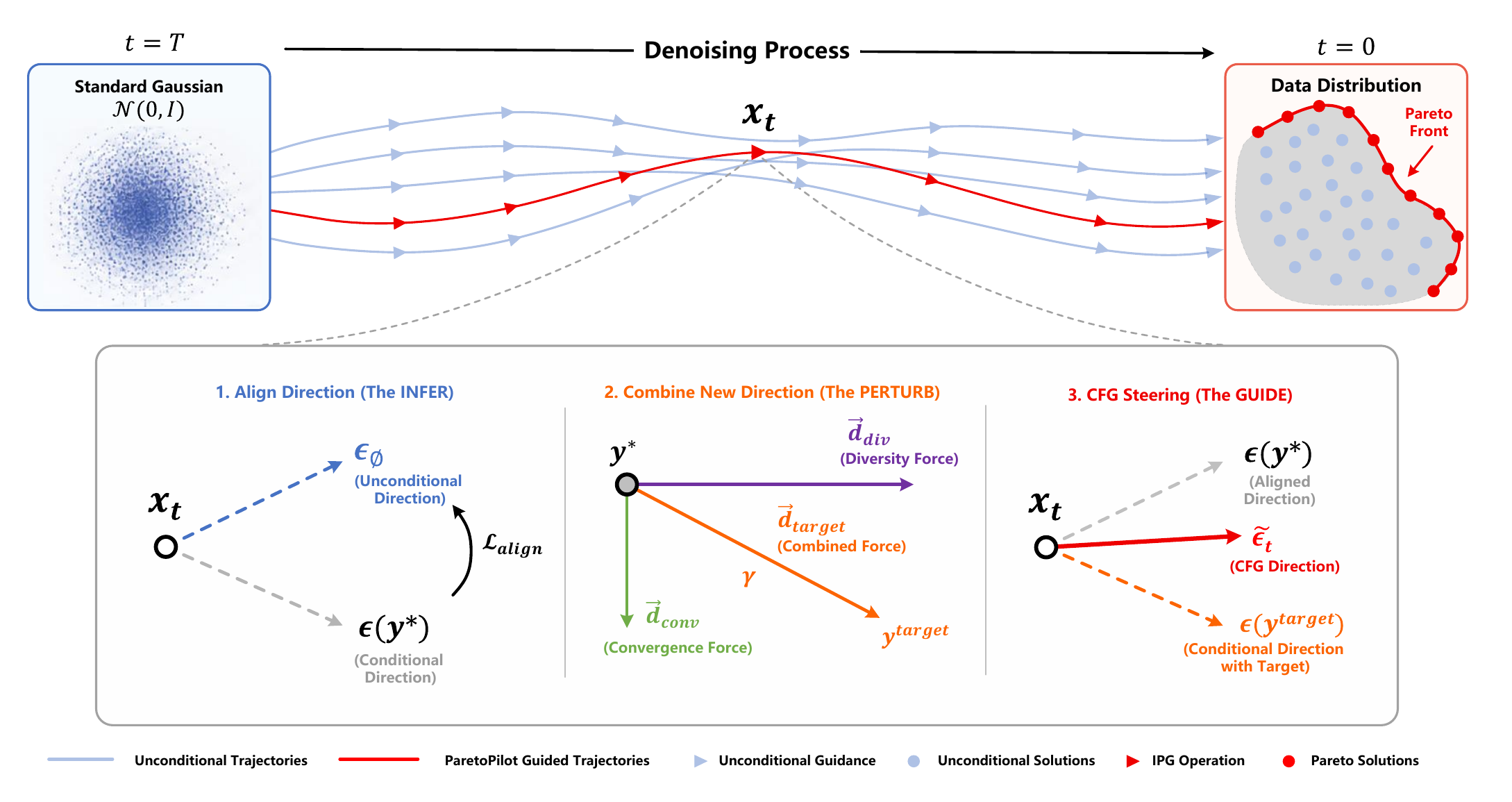}
    \caption{\textbf{Overview of the ParetoPilot framework featuring the IPG engine.}  After every $V$ unconditional denoising steps, IPG adjusts the denoising direction. \textbf{1. INFER:} It locates the instantaneous direction anchor $y^*$ by aligning the conditional noise prediction $\epsilon(y^*)$ with the unconditional direction $\epsilon_{\emptyset}$. \textbf{2.PERTURB:} It derives a perturbed coordinate $y^{target}$ by shifting $y^*$ along a synthesized direction $\vec{d}_{target}$ that mathematically decouples the convergence and diversity forces. \textbf{3. GUIDE:} It utilizes the perturbed target $y^{target}$ within the CFG to steer the final diffusion trajectory $\tilde{\epsilon}_t$ toward optimal designs without external surrogates.}
    \label{fig:teaser}
\end{figure*}

\subsection{Overview of ParetoPilot}
\label{sec:method_overview}

ParetoPilot addresses the above challenges by modifying only the conditioning targets and guidance strength in the standard CFG formulation. During the reverse diffusion process, unconditional denoising and IPG updates are interleaved: the former preserves sample plausibility, while the latter periodically steers the batch of candidates toward the Pareto front. Specifically, ParetoPilot performs an IPG update every $V$ denoising steps. Each update consists of three stages: \textbf{Infer}, \textbf{Perturb}, and \textbf{Guide}. We detail these stages below.

\subsection{INFER: Implicit Manifold Anchoring and Noise Filtering}
\label{sec:method_infer}
To guide a noisy sample $x_t$, we must first locate its corresponding instantaneous objective direction $y_t^*$. Instead of relying on an external surrogate regressor, we implicitly anchor $y_t^*$ by leveraging the internal knowledge of the model. Specifically, the natural condition direction $y_t^*$ is the configuration where the conditional noise prediction perfectly aligns with the unconditional noise prediction, resulting in zero guidance momentum. We define the alignment loss as:
\begin{equation}
    \mathcal{L}_{align}(y_t) = \left\| \epsilon_\theta(x_t, y_t, t) - \epsilon_\theta(x_t, \emptyset, t) \right\|_2^2
\end{equation}
Minimizing this loss via gradient descent yields the instantaneous anchor direction. However, the noise predictions $\epsilon_\theta$ at large timesteps $t$ are highly stochastic, causing significant high-frequency jitter across the diffusion steps if we optimize $y_t$ naively.

To filter out this inherent stochasticity, we leverage the internal momentum mechanics of an \emph{un-reset Adam optimizer}. Rather than re-initializing the optimizer state at each probing step $t$, we maintain a global Adam optimizer across the entire diffusion process. During the gradient updates to minimize the alignment loss, the optimizer naturally computes and preserves the exponential moving average of the past gradients. By carrying these internal momentum states forward across the diffusion timesteps, the optimizer intrinsically acts as a low-pass filter. This mechanism smooths out the high-frequency jitter of the diffusion noise, enabling a stable trajectory for $y_t^*$ as it gradually approaches the optimal Pareto manifold.

\subsection{PERTURB: Orthogonal Fluid Dynamics}
\label{sec:method_perturb}
Once the anchor $y_t^*$ is obtained, we synthesize a multi-objective perturbation vector in the latent objective space. To simultaneously guarantee rigorous convergence and uniform Pareto front coverage without mutual interference, we construct an orthogonal fluid dynamics system.

Before applying the perturbations, we introduce a warmup phase comprising the initial steps of the reverse diffusion process where $t > T - T_{warmup}$. This warmup phase serves three critical purposes. First, it provides sufficient time for the alignment loss to stabilize, ensuring that the inferred anchor direction $y_t^*$ is accurate before any perturbation is introduced. Second, applying guidance too early in the high-noise regime is empirically unnecessary and introduces redundant computational overhead. Third, aggressive perturbations in the early high-noise stage may push samples away from the valid denoising trajectory and cause manifold collapse. Consequently, we strictly confine the kinematic perturbations to the steps where $t \le T - T_{warmup}$. For this active phase, we introduce a normalized scheduling variable $\tau_t = t / (T - T_{warmup}) \in [0, 1]$ that linearly decays as the denoising process approaches $t=0$.

\textbf{Prior-Preserving Parallel Gravity.} To drive the samples toward the optimal regions, we apply a unified parallel gravity field across all particles:
\begin{equation}
    \vec{d}_{conv} = \frac{-\mathbf{1}}{\|\mathbf{-1}\|_2}
\end{equation}
Geometrically, as all solutions descend in parallel along the constant vector, their relative lateral distances are strictly preserved. This parallel exploitation mathematically retains the inherent prior diversity generated by the initial unconditional diffusion distribution.

\textbf{Edgeness-Aware Repulsion.} Due to the isotropic nature of the initial Gaussian noise distribution, the reverse mapped solutions might not naturally spread wide enough to cover the extreme regions of the Pareto front. However, certain solutions will inherently land closer to these extrema. To actively expand the boundaries and ensure complete coverage, we must apply an additional outward thrust to these marginal solutions. We achieve this by introducing an \emph{edgeness} multiplier. We min-max normalize the coordinates across the batch to obtain $\tilde{y}_{i} \in [0,1]^m$. The edgeness factor for the $i$-th particle is defined as:
\begin{equation}
    \mathcal{E}_i = \exp(-\frac{1}{2} \min_k \tilde{y}_{i,k})
\end{equation}
Solutions located near the boundaries receive relatively stronger repulsive momentum, effectively pushing them outward to discover extreme trade-off designs. The raw repulsive force exerted on the $i$-th particle is then computed as:
\begin{equation}
    \vec{F}_{rep, i} = \sum_{j \neq i} \frac{y_i^* - y_j^*}{\|y_i^* - y_j^*\|_2^2 + \delta}
\end{equation}
where $\delta$ is a small constant for numerical stability, and the restriction $j \neq i$ prevents self-repulsion.

\textbf{Gram-Schmidt Orthogonalization.} A fundamental flaw in existing diversity-promoting techniques is that repulsive forces inevitably produce components that oppose the direction of convergence, thereby slowing down or degrading exploitation. To resolve this mathematically, we project the repulsive force onto the orthogonal complement of the gravity field using Gram-Schmidt orthogonalization:
\begin{equation}
    \vec{F}_{\perp, i} = \vec{F}_{rep, i} - \left( \vec{F}_{rep, i} \cdot \vec{d}_{conv} \right) \vec{d}_{conv}
\end{equation}
We then normalize this orthogonal vector to obtain the pure diversity direction $\vec{d}_{div, i} = \vec{F}_{\perp, i} / \|\vec{F}_{\perp, i}\|_2$. This strict orthogonalization ensures that $\vec{d}_{div, i} \perp \vec{d}_{conv}$, mathematically decoupling exploration from exploitation.

\textbf{Dynamic Alpha Annealing.} Diffusion models inherently exhibit a coarse-to-fine generation trajectory. Correspondingly, we design a temporal annealing schedule for the kinematic synthesis. The blending weight $\alpha_t$ smoothly transitions from $\alpha_{start}$ to $\alpha_{end}$ based on the normalized timestep $\tau_t$:
\begin{equation}
    \alpha_t = \alpha_{end} + (\alpha_{start} - \alpha_{end}) \cdot \tau_t
\end{equation}
The final synthesized perturbation direction is a normalized combination of convergence and edge-aware diversity:
\begin{equation}
    \vec{d}_{target, i} = \frac{\alpha_t \vec{d}_{conv} + (1-\alpha_t) \mathcal{E}_i \vec{d}_{div, i}}{\left\| \alpha_t \vec{d}_{conv} + (1-\alpha_t) \mathcal{E}_i \vec{d}_{div, i} \right\|_2}
\end{equation}
Finally, the target objective direction is obtained by advancing a fixed linear radius $\gamma$:
\begin{equation}
    y_{t}^{target} = y_t^* + \gamma \cdot \vec{d}_{target, i}
\end{equation}

\subsection{GUIDE: CFG Extrapolation}
\label{sec:method_guide}
With the dynamically perturbed target direction $y_t^{target}$, we directly inject it back into the standard CFG formulation to guide the diffusion generation. To complement the temporal dynamics of the perturbation, we also linearly interpolate the CFG scale $w_t$ alongside the denoising process:
\begin{equation}
    w_t = w_{end} + (w_{start} - w_{end}) \cdot \tau_t
\end{equation}
The modified noise prediction used to update the sample $x_t$ is computed as:
\begin{equation}
    \tilde{\epsilon}_t = \epsilon_\theta(x_t, \emptyset, t) + w_t \cdot \left( \epsilon_\theta(x_t, y_t^{target}, t) - \epsilon_\theta(x_t, \emptyset, t) \right)
\end{equation}
By keeping $\gamma$ as a small and fixed kinematic step size, we prevent severe non-linear extrapolation errors in the neural network predictions. The extrapolation intensity is instead controlled safely and independently by the CFG weight $w_t$, ensuring a stable and effective condition-guided generation.

\begin{table*}[t]
\centering
\caption{\textbf{Quantitative Results on Off-MOO-Bench.} We report the Average Rank (lower is better) and the Percentage Improvement of Hypervolume (higher is better) across four benchmark domains. The best results across all methods are highlighted in \textbf{bold}, while the second-best results are \underline{underlined}.}
\label{tab:main_results}
\resizebox{\textwidth}{!}{
\begin{tabular}{l|ccccc|ccccc}
\toprule
\multirow{2}{*}{\textbf{Method}} & \multicolumn{5}{c|}{\textbf{Average Rank ($\downarrow$)}} & \multicolumn{5}{c}{\textbf{Percentage Improvement of HV ($\uparrow$)}} \\
\cmidrule(lr){2-6} \cmidrule(lr){7-11}
 & Synthetic & MO-NAS & SciDesign & RE Suite & \textbf{Overall} & Synthetic & MO-NAS & SciDesign & RE Suite & \textbf{Overall} \\
\midrule
\multicolumn{11}{c}{\textit{Surrogate-based Forward Methods}} \\
\midrule
MultipleModels-IOM & 7.81 & 6.72 & 5.67 & 6.67 & \underline{6.98} & 29.73\% & -4.03\% & 19.71\% & 11.12\% & \textbf{12.10\%} \\
MultipleModels-Vanilla & \underline{4.31} & 10.17 & 12.67 & 6.67 & 7.50 & \textbf{32.26\%} & -11.21\% & -13.20\% & 10.35\% & 8.27\% \\
MultipleModels-ICT & 7.19 & 9.00 & 8.67 & 6.00 & 7.56 & 29.49\% & -9.51\% & 15.53\% & \underline{11.52\%} & 10.00\% \\
MultiHead-Vanilla & \textbf{4.25} & 11.44 & 15.00 & \underline{5.13} & 7.62 & 29.97\% & -12.07\% & 4.95\% & 10.96\% & 8.56\% \\
End2End-PcGrad & 6.06 & 11.67 & 7.67 & 5.73 & 8.00 & 30.99\% & -12.31\% & 11.64\% & 11.14\% & 9.16\% \\
End2End-Vanilla & 5.00 & 10.67 & 12.67 & 7.73 & 8.19 & \underline{32.09\%} & -11.14\% & -12.63\% & 9.34\% & 7.98\% \\
MultipleModels-TriMentoring & 9.50 & 10.11 & 7.33 & \textbf{5.00} & 8.29 & 27.61\% & -9.92\% & 17.78\% & \textbf{11.56\%} & 9.42\% \\
MultiHead-PcGrad & 8.12 & 11.11 & 10.33 & 6.20 & 8.73 & 27.31\% & -11.01\% & 11.22\% & 11.07\% & 8.43\% \\
MultipleModels-COM & 11.06 & 5.00 & 5.67 & 12.80 & 9.15 & 19.12\% & \underline{-1.48\%} & 18.60\% & 4.27\% & 7.68\% \\
MultipleModels-RoMA & 10.19 & 10.72 & 5.67 & 9.87 & 10.02 & 24.63\% & -12.72\% & 19.12\% & 9.47\% & 7.01\% \\
MultiHead-GradNorm & 10.06 & 12.33 & 13.00 & 9.00 & 10.71 & 19.56\% & -15.46\% & -1.79\% & 6.50\% & 2.44\% \\
End2End-GradNorm & 10.06 & 11.61 & 8.00 & 11.00 & 10.75 & 27.82\% & -13.90\% & 12.47\% & 9.02\% & 7.07\% \\
\midrule
\multicolumn{11}{c}{\textit{Surrogate-based Generative Inverse Methods}} \\
\midrule
PGD-MOO & 11.81 & \textbf{2.00} & \textbf{2.00} & 10.93 & 7.60 & 4.38\% & \textbf{0.37\%} & \textbf{28.63\%} & 6.60\% & 5.04\% \\
ParetoFlow & 11.81 & 5.72 & 11.67 & 8.47 & 8.73 & 27.58\% & -3.30\% & -3.03\% & 8.95\% & 9.75\% \\
\midrule
\multicolumn{11}{c}{\textit{Zero-Surrogate Generative Inverse Methods}} \\
\midrule
PCD & 9.69 & \underline{3.67} & \underline{4.33} & 12.53 & 8.12 & 7.54\% & -2.18\% & \underline{25.11\%} & 2.72\% & 3.92\% \\
\textbf{ParetoPilot (Ours)} & 8.12 & 4.06 & 5.67 & 9.07 & \textbf{6.85} & 27.42\% & -2.36\% & 25.01\% & 6.63\% & \underline{10.98\%} \\
\bottomrule
\end{tabular}
}
\end{table*}

\section{Experiments}
\label{sec:experiments}

\subsection{Experimental Setup}
\label{sec:setup}

\textbf{Benchmark Tasks.} 
We evaluate our method across $51$ continuous offline MOO tasks from the Off-MOO-Bench platform \cite{xue2024offline}. The benchmark comprises four distinct domains. The Synthetic Functions domain includes $15$ tasks (DTLZ1 to 7, OmniTest, VLMOP1 to 3, ZDT1 to 4, and ZDT6). The Multi-Objective Neural Architecture Search (MO-NAS) domain consists of $18$ tasks operating in continuous logit spaces (C10MOP1 to 9, and IN1KMOP1 to 9). The Scientific Design (SciDesign) domain contains $3$ high-dimensional optimization tasks (Molecule, Regex, and ZINC). Finally, the Real-world Engineering (RE) Suite includes $15$ industrial design problems (RE21 to 25, RE31 to 37, RE41, RE42, and RE61).

\textbf{Baseline Methods.} 
We compare ParetoPilot against $15$ state-of-the-art offline MOO algorithms. The surrogate-based forward methods include End-to-End architectures (Vanilla, PcGrad, GradNorm), Multi-Head networks (Vanilla, PcGrad, GradNorm), and Multiple Independent Models equipped with various robust adaptation strategies (Vanilla, COMs, RoMA, IOM, ICT, Tri-Mentoring). The generative inverse methods include ParetoFlow, PGD-MOO and PCD.

\textbf{Evaluation Metrics.} 
We assess overall algorithmic stability by computing the \emph{Average Rank of Hypervolume (HV)} across all tasks. Additionally, we calculate the \emph{Percentage Improvement of HV} relative to the best solutions in the offline dataset to explicitly quantify the extrapolation capability in discovering novel superior designs.

\textbf{Implementation Details.} 
We use a conditional diffusion model as the backbone for ParetoPilot. The IPG engine operates at an interval of $V=10$ and performs $k=5$ probing steps per interval with a learning rate of $0.001$. The warmup duration is set to $T_{warmup}=500$ timesteps. The dynamic annealing parameters $\alpha_t$ and $w_t$ transition from $0.6$ to $0.45$ and from $1.8$ to $2.8$, respectively, and the perturbation step size is fixed at $\gamma=0.45$. We conduct all experiments across $4$ independent random seeds. All model training and evaluation procedures are executed on a single workstation equipped with an RTX 3090 GPU.

\subsection{Results}
\label{sec:main_results}

The results reveal clear domain-dependent behaviors among existing methods. Approaches that rely on objective surrogate models, including forward optimization methods and ParetoFlow, perform competitively on continuous Synthetic and RE tasks. In these domains, the design spaces are relatively smooth and most generated coordinates correspond to valid objective evaluations, making surrogate-guided extrapolation effective. However, their performance degrades substantially on more structured or discrete tasks such as MO-NAS and SciDesign, where invalid or out-of-distribution designs are more likely to appear. In such settings, neural surrogate models may provide unreliable scores outside the data distribution, which can mislead optimization and result in negative HV improvements.

PGD-MOO alleviates part of this issue by replacing absolute objective prediction with an external preference classifier. This relative comparison strategy is more robust to noisy objective values and leads to strong results on SciDesign and MO-NAS. However, the preference-guided formulation also tends to be conservative, since pairwise dominance signals provide limited direction for aggressive extrapolation. Consequently, PGD-MOO performs less competitively on Synthetic and RE tasks, where extrapolating beyond the observed objective range is often beneficial. In addition, training the auxiliary preference classifier introduces non-negligible computational overhead.

The recent zero-surrogate method PCD improves cross-domain robustness by avoiding external neural proxies. However, it still relies on static conditioning points heuristically constructed from the offline dataset. Since these targets are fixed before sampling, they are not guaranteed to fall within the controllable region of the learned conditional diffusion model. Consequently, overly aggressive conditions may push generation outside the range where the model can produce plausible samples. This partly explains PCD's weak performance on the Synthetic tasks, where many generated solutions move beyond the reasonable data range. Moreover, its Gaussian-perturbation strategy encourages only local diversity around static targets, which may be insufficient to cover the full Pareto front.

In contrast, ParetoPilot achieves more consistent performance across domains. Since it operates entirely at inference time without external proxies, it avoids surrogate-induced evaluation errors and auxiliary proxy-training overhead. During sampling, unconditional denoising helps keep generated candidates close to the learned data manifold, while the IPG engine dynamically adjusts conditioning targets to balance convergence toward the Pareto front and diversity among solutions. As a result, ParetoPilot obtains the best overall rank among all methods and achieves an overall HV improvement of $10.98\%$.

\begin{figure*}[t]
    \centering
    \includegraphics[width=\textwidth]{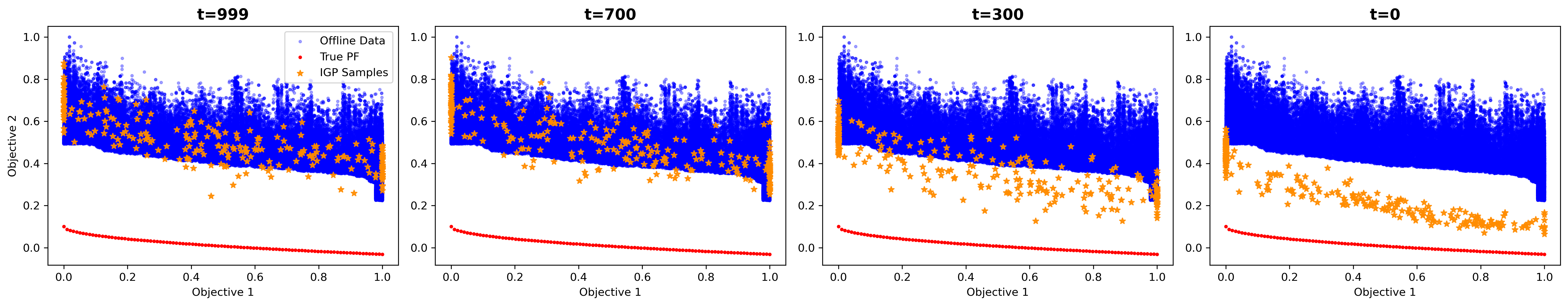}
    \caption{\textbf{Visualization of the dynamic generation trajectory of ParetoPilot on ZDT1.} At $t$=999, the generation starts with the initial noisy distribution. At $t$=700, the process is still within the warm-up phase without perturb and guide intervention, leaving the solution distribution similar to the initial state. At $t$=300, the active IPG engine dynamically steers the samples out of the offline data distribution towards the Pareto optimal region. Finally, at $t$=0, the generated samples achieve highly robust convergence and diversity to the Pareto optimal region. 
    }
    \label{fig:trajectory}
\end{figure*}

\subsection{Ablation Studies}
\label{sec:ablation}

\begin{table*}[t]
\centering
\caption{\textbf{Ablation study of the IPG engine.} We validate the effectiveness of key components by comparing the full ParetoPilot against three variants: disabling the historical momentum (\textit{w/o Smoothing}), removing the boundary amplification factor (\textit{w/o Edgeness}), and directly blending forces without Gram-Schmidt projection (\textit{w/o Orthogonalization}).}\label{tab:ablation}
\resizebox{\textwidth}{!}{
\begin{tabular}{l|ccccc|ccccc}
\toprule
\multirow{2}{*}{\textbf{Method Variant}} & \multicolumn{5}{c|}{\textbf{Average Rank ($\downarrow$)}} & \multicolumn{5}{c}{\textbf{Percentage Improvement of HV ($\uparrow$)}} \\
\cmidrule(lr){2-6} \cmidrule(lr){7-11}
 & Synthetic & MO-NAS & SciDesign & RE Suite & \textbf{Overall} & Synthetic & MO-NAS & SciDesign & RE Suite & \textbf{Overall} \\
\midrule
w/o Smoothing & 2.94 & 2.44 & 2.50 & 3.60 & 2.94 & 26.16\% & -2.15\% & \textbf{27.15\%} & -0.94\% & 8.24\% \\
w/o Edgeness & 2.62 & 2.39 & 3.50 & 2.33 & 2.49 & 26.94\% & -2.11\% & 24.05\% & 6.56\% & 10.58\% \\
w/o Orthogonalization & \textbf{2.12} & \textbf{2.78} & \textbf{2.00} & 2.20 & 2.37 & 27.23\% & -2.23\% & 25.52\% & 5.21\% & 10.29\% \\
\midrule
ParetoPilot (Ours Full) & 2.31 & 2.33 & \textbf{2.00} & \textbf{1.87} & \textbf{2.17} & \textbf{27.42\%} & \textbf{-2.36\%} & 25.01\% & \textbf{6.63\%} & \textbf{10.98\%} \\
\bottomrule
\end{tabular}
}
\end{table*}

To validate the efficacy of each kinematic and optimization component within the proposed ParetoPilot framework, we construct three specific ablation variants by systematically isolating key mechanisms. First, we design the \emph{w/o Smoothing} variant by resetting the Adam optimizer states at the beginning of every probing step to disable the historical exponential moving average momentum. Second, we create the \emph{w/o Edgeness} variant by removing the boundary amplification factor, compelling the system to rely solely on uniform distance-based repulsion. Third, we evaluate the \emph{w/o Orthogonalization} variant by directly blending the raw repulsive force with the convergence direction without the Gram-Schmidt projection. 

The empirical results consistently confirm the importance of each proposed component. The \textbf{w/o Smoothing} variant suffers a significant performance drop, plunging to an overall average rank of 2.94 and a drastically reduced hypervolume improvement of 8.24\%. Notably, it yields a severe negative improvement of $-0.94\%$ in the RE Suite, suggesting that unfiltered high-frequency noise can cause trajectory oscillations that push samples off the valid data manifold. Similarly, the \textbf{w/o Edgeness} variant experiences a noticeable performance drop to an overall rank of 2.49, confirming that particles fail to efficiently discover extreme trade-off designs without the exponentially amplified outward thrust at the margins. Finally, the \textbf{w/o Orthogonalization} variant degrades the overall hypervolume improvement to 10.29\% with an average rank of 2.37. This decline empirically supports our geometric hypothesis that raw repulsion inherently contains components opposing the gravity vector. Without strict orthogonal projection, these non-orthogonal forces mutually neutralize the convergence momentum, demonstrating that the Gram-Schmidt mechanism is important for decoupling exploration from exploitation and maintaining steady optimization speed.

\section{Conclusion}
\label{sec:conclusion}

In this work, we proposed ParetoPilot as a zero-surrogate generative framework for offline multi-objective optimization. By avoiding the reliance on auxiliary proxy models for guidance, our approach seamlessly integrates with pre-trained conditional diffusion models. This design alleviates the need for direct raw data exposure and aligns well with practical downstream deployment scenarios where access to the original training dataset is restricted due to proprietary or storage constraints. Through the introduced IPG engine, we dynamically navigated the latent condition space to effectively balance convergence toward the optimal boundary and diversity across the objective space. Extensive empirical evaluations demonstrated that this plug-and-play methodology consistently achieves robust extrapolation and competitive performance across complex search spaces compared to existing surrogate-reliant baselines.

Despite these advantages, our framework possesses certain limitations that offer valuable opportunities for future research. The overall generation quality inherently depends on the capacity of the foundational diffusion model to accurately capture the initial offline data distribution. If the base model fails to model the underlying manifold appropriately, the latent kinematic navigation will be consequently impaired. In addition, the alignment loss and gradient probing steps introduce extra inference latency. Furthermore, integrating explicit user-defined constraints directly into the perturbation step presents an exciting avenue for expanding the applicability of zero-surrogate optimization in highly regulated industrial environments.

\bibliography{aaai2027}


\section{Appendix A: Theoretical Derivations}
\label{app:theory}

This section provides the theoretical justification for the three key operations in the IPG engine. We first clarify why the Infer step can recover a self-consistent objective anchor from a frozen conditional diffusion model. We then prove that the orthogonal perturbation preserves the convergence component exactly. Finally, we show that the un-reset Adam momentum acts as a temporal low-pass filter during diffusion sampling.

\subsection{A.1 Self-Consistent Condition Recovery in the Infer Step}
\label{app:infer_derivation}

Let $\epsilon_\theta(x_t,\emptyset,t)$ denote the unconditional noise prediction at timestep $t$, and $\epsilon_\theta(x_t,y,t)$ denote the conditional prediction under an objective-space condition $y$. In standard CFG, the guided prediction is
\begin{equation}
    \tilde{\epsilon}_\theta(x_t,y,t)
    =
    \epsilon_\theta(x_t,\emptyset,t)
    +
    w\left(
    \epsilon_\theta(x_t,y,t)
    -
    \epsilon_\theta(x_t,\emptyset,t)
    \right),
\end{equation}
where $w$ is the guidance scale. The term
\begin{equation}
    \Delta_\theta(x_t,y,t)
    =
    \epsilon_\theta(x_t,y,t)
    -
    \epsilon_\theta(x_t,\emptyset,t)
\end{equation}
is therefore the additional steering momentum induced by the condition $y$.

The key observation is that the unconditional prediction $\epsilon_\theta(x_t,\emptyset,t)$ represents the model's natural denoising direction toward the learned data manifold, without any external preference imposed. Thus, if a condition $y_t^*$ satisfies
\begin{equation}
    \epsilon_\theta(x_t,y_t^*,t)
    \approx
    \epsilon_\theta(x_t,\emptyset,t),
\end{equation}
then applying CFG with $y_t^*$ introduces nearly zero extra steering:
\begin{equation}
    \Delta_\theta(x_t,y_t^*,t) \approx 0.
\end{equation}
In this case, $y_t^*$ can be interpreted as the \emph{self-consistent objective anchor} of the current noisy sample: it is the condition under which the conditional model agrees with the unconditional manifold-preserving denoising direction.

Accordingly, ParetoPilot recovers this anchor by solving the following inverse alignment problem:
\begin{equation}
    y_t^*
    =
    \arg\min_{y}
    \mathcal{L}_{\mathrm{align}}(y),
    \quad
    \mathcal{L}_{\mathrm{align}}(y)
    =
    \left\|
    \epsilon_\theta(x_t,y,t)
    -
    \epsilon_\theta(x_t,\emptyset,t)
    \right\|_2^2.
\end{equation}
This optimization does not require any external surrogate model. Instead, it inverts the conditional prior already stored in the frozen diffusion model and identifies the objective-space coordinate that is most compatible with the current sample and its manifold-preserving denoising direction.

After $y_t^*$ is obtained, ParetoPilot does not directly use it as the final target. Instead, $y_t^*$ serves as a local anchor in the objective space, from which IPG constructs a perturbed target that improves convergence and diversity.

\subsection{A.2 Convergence-Preserving Orthogonal Perturbation}
\label{app:orthogonal_derivation}

After the anchor $y_t^*$ is inferred for each sample in the batch, IPG constructs two objective-space forces: a convergence force $\mathbf{g}_i$ that moves the $i$-th solution toward the Pareto front, and a diversity force $\mathbf{r}_i$ that repels it from other solutions. A naive update
\begin{equation}
    \mathbf{d}_i = \mathbf{g}_i + \alpha_t \mathbf{r}_i
\end{equation}
does not guarantee convergence, because the diversity force may contain a component opposite to $\mathbf{g}_i$. If $\mathbf{r}_i^\top \mathbf{g}_i < 0$, the repulsion partially cancels the convergence momentum.

To remove this interference, IPG projects $\mathbf{r}_i$ onto the subspace orthogonal to $\mathbf{g}_i$:
\begin{equation}
    \mathbf{r}_i^\perp
    =
    \mathbf{r}_i
    -
    \frac{\mathbf{r}_i^\top \mathbf{g}_i}
    {\|\mathbf{g}_i\|_2^2 + \delta}
    \mathbf{g}_i,
\end{equation}
where $\delta>0$ is a small constant for numerical stability. The final perturbed direction is
\begin{equation}
    \mathbf{d}_i^*
    =
    \mathbf{g}_i + \alpha_t \mathbf{r}_i^\perp .
\end{equation}

\textbf{Proposition.}
If $\mathbf{g}_i \neq \mathbf{0}$ and $\delta \rightarrow 0$, the perturbed direction $\mathbf{d}_i^*$ preserves the exact convergence projection of $\mathbf{g}_i$.

\textbf{Proof.}
By construction,
\begin{equation}
\begin{aligned}
    (\mathbf{r}_i^\perp)^\top \mathbf{g}_i
    &=
    \left(
    \mathbf{r}_i
    -
    \frac{\mathbf{r}_i^\top \mathbf{g}_i}
    {\|\mathbf{g}_i\|_2^2}
    \mathbf{g}_i
    \right)^\top \mathbf{g}_i \\
    &=
    \mathbf{r}_i^\top \mathbf{g}_i
    -
    \frac{\mathbf{r}_i^\top \mathbf{g}_i}
    {\|\mathbf{g}_i\|_2^2}
    \|\mathbf{g}_i\|_2^2 \\
    &= 0.
\end{aligned}
\end{equation}
Therefore,
\begin{equation}
\begin{aligned}
    (\mathbf{d}_i^*)^\top \mathbf{g}_i
    &=
    \left(
    \mathbf{g}_i + \alpha_t \mathbf{r}_i^\perp
    \right)^\top \mathbf{g}_i \\
    &=
    \|\mathbf{g}_i\|_2^2
    +
    \alpha_t
    (\mathbf{r}_i^\perp)^\top \mathbf{g}_i \\
    &=
    \|\mathbf{g}_i\|_2^2 .
\end{aligned}
\end{equation}
Thus, the diversity force contributes zero projection along the convergence axis and cannot reduce the convergence component. This proves that orthogonalization decouples diversity enhancement from convergence acceleration.

\subsection{A.3 Un-reset Adam Momentum as a Temporal Low-Pass Filter}
\label{app:adam_derivation}

During Infer, the anchor $y_t^*$ is obtained by minimizing $\mathcal{L}_{\mathrm{align}}(y)$ across diffusion timesteps. Because the noisy state $x_t$ changes over time and diffusion sampling is stochastic, the gradient sequence
\begin{equation}
    \mathbf{q}_t = \nabla_y \mathcal{L}_{\mathrm{align}}(y_t)
\end{equation}
contains high-frequency fluctuations. If each timestep uses a freshly initialized optimizer, the update direction is dominated by the instantaneous noisy gradient $\mathbf{q}_t$.

ParetoPilot instead uses an un-reset Adam optimizer. Its first-order moment follows
\begin{equation}
    \mathbf{m}_t
    =
    \beta_1 \mathbf{m}_{t-1}
    +
    (1-\beta_1)\mathbf{q}_t .
\end{equation}
Unrolling the recurrence gives
\begin{equation}
    \mathbf{m}_t
    =
    (1-\beta_1)
    \sum_{k=0}^{t}
    \beta_1^k
    \mathbf{q}_{t-k},
\end{equation}
assuming $\mathbf{m}_0=\mathbf{0}$. Therefore, $\mathbf{m}_t$ is an exponentially weighted moving average of historical gradients.

For a scalar gradient sequence, the corresponding transfer function of this exponential moving average is
\begin{equation}
    H(z)
    =
    \frac{1-\beta_1}{1-\beta_1 z^{-1}}.
\end{equation}
Evaluating it on the unit circle $z=e^{j\omega}$ yields the frequency response
\begin{equation}
    |H(e^{j\omega})|
    =
    \frac{1-\beta_1}
    {\sqrt{1+\beta_1^2-2\beta_1\cos\omega}} .
\end{equation}
This response is largest at low frequency:
\begin{equation}
    |H(e^{j0})| = 1,
\end{equation}
and smallest at the highest frequency:
\begin{equation}
    |H(e^{j\pi})|
    =
    \frac{1-\beta_1}{1+\beta_1}.
\end{equation}
For the commonly used $\beta_1=0.9$, the highest-frequency component is attenuated to approximately $0.0526$ of its original magnitude.

Thus, the un-reset first-order Adam momentum acts as a temporal low-pass filter: it preserves slowly varying, consistent optimization trends while suppressing timestep-to-timestep jitter. This explains why the w/o Smoothing variant exhibits unstable trajectories and degraded hypervolume in our ablation study.

\section{Appendix B: Detailed Experimental Setup and Model Architecture}
\label{app:setup}

This section provides the implementation details of ParetoPilot, including the conditional diffusion backbone, the DDPM training protocol, the inference-time IPG configuration, and the baseline settings. Unless otherwise specified, all experiments use the same configuration across benchmark tasks without task-specific tuning.

\subsection{B.1 Conditional Diffusion Backbone}
\label{app:model_architecture}

ParetoPilot is built upon a lightweight conditional DDPM backbone trained with CFG. The model takes a noisy design vector $x_t \in \mathbb{R}^{d_x}$, a diffusion timestep $t$, and an optional objective-space condition $y \in \mathbb{R}^{m}$ as inputs, where $d_x$ is the task-specific design dimension and $m$ is the number of objectives.

The denoiser contains an input projection layer, a sinusoidal timestep encoder, an objective-condition encoder, two AdaLN-modulated MLP blocks, and a final output projection. The default hidden dimension of the denoising backbone is 256. In our main experiments, the hidden dimension of the condition encoder is set to 32, and its final projection maps the encoded objective condition to the 256-dimensional modulation space.

\begin{table}[h]
\centering
\caption{\textbf{Architecture of the conditional diffusion backbone.} Here $d_x$ denotes the task-specific design dimension and $m$ denotes the number of objectives. The right column uses automatic line wrapping for readability.}
\label{tab:model_architecture}
\resizebox{\linewidth}{!}{
\begin{tabular}{l|p{0.62\linewidth}}
\toprule
\textbf{Component} & \textbf{Configuration} \\
\midrule
Input projection & Linear$(d_x, 256)$ maps the task-specific design vector to the hidden space. \\
Timestep encoding & Sinusoidal positional encoding with dimension 256. \\
Timestep MLP & Linear$(256,256)$ + ReLU + Linear$(256,256)$. \\
Condition encoder & Linear$(m,32)$ + ReLU + Linear$(32,32)$ + ReLU + Linear$(32,256)$. \\
Null condition embedding & Learnable unconditional embedding in $\mathbb{R}^{256}$, used when the condition is dropped or absent. \\
AdaLN block 1 & Linear$(256,256)$ + ReLU + non-affine LayerNorm, modulated by condition-dependent scale and shift. \\
AdaLN block 2 & Linear$(256,256)$ + ReLU + non-affine LayerNorm, modulated by condition-dependent scale and shift. \\
Output projection & Linear$(256,d_x)$ predicts the diffusion noise in the original design dimension. \\
Activation & ReLU is used in both timestep/condition encoders and AdaLN blocks. \\
Initialization & The AdaLN scale-shift projection and the final output layer are zero-initialized for stable conditional training. \\
Condition dropout & 0.1 during training for CFG. \\
\bottomrule
\end{tabular}
}
\end{table}

Each AdaLN block transforms the hidden representation and then modulates it using the joint timestep-condition signal. Given hidden state $h$ and modulation signal $c$, the block computes
\begin{equation}
    \mathrm{AdaLN}(h,c)
    =
    \mathrm{LN}(\phi(W h)) \odot (1+s(c)) + b(c),
\end{equation}
where $\phi$ is ReLU, $\mathrm{LN}$ is LayerNorm without affine parameters, and $s(c),b(c)$ are predicted from the modulation signal. The scale-shift projection is zero-initialized so that each AdaLN block initially behaves like a standard normalized MLP block, improving training stability.

\subsection{B.2 Diffusion Training Protocol}
\label{app:diffusion_training}

We train the conditional diffusion backbone with the standard DDPM noise-prediction objective. Given a normalized design vector $x_0$, timestep $t$, and Gaussian noise $\epsilon \sim \mathcal{N}(0,I)$, the forward noising process is
\begin{equation}
    x_t
    =
    \sqrt{\bar{\alpha}_t}x_0
    +
    \sqrt{1-\bar{\alpha}_t}\epsilon.
\end{equation}
The model is optimized to predict the injected noise:
\begin{equation}
    \mathcal{L}_{\mathrm{DDPM}}
    =
    \mathbb{E}_{x_0,t,\epsilon}
    \left[
    \left\|
    \epsilon -
    \epsilon_\theta(x_t,t,y)
    \right\|_2^2
    \right].
\end{equation}

The training and diffusion hyperparameters are summarized in Table~\ref{tab:training_hyperparameters}. All design vectors are normalized by the benchmark preprocessing pipeline and rescaled to $[-1,1]$ before being fed into the diffusion model.

\begin{table}[h]
\centering
\caption{\textbf{Training hyperparameters for the conditional diffusion backbone.}}
\label{tab:training_hyperparameters}
\resizebox{\linewidth}{!}{
\begin{tabular}{l|p{0.56\linewidth}}
\toprule
\textbf{Hyperparameter} & \textbf{Value} \\
\midrule
Diffusion steps & 1000 \\
Noise schedule & Linear schedule from $\beta_{\mathrm{start}}=1\times10^{-4}$ to $\beta_{\mathrm{end}}=2\times10^{-2}$. \\
Training epochs & 200 \\
Optimizer & AdamW \\
Learning rate & $5\times10^{-4}$ \\
Gradient clipping & 1.0 \\
Condition dropout probability & 0.1 \\
EMA decay for model weights & 0.99 \\
EMA warmup steps & 2000 \\
Validation ratio & 0.9, following the dataloader split used in our implementation. \\
\bottomrule
\end{tabular}
}
\end{table}

\section{Appendix C: Hyperparameter Sensitivity Analysis}
\label{sec:appendix_sensitivity}

In this section, we comprehensively investigate the sensitivity of the proposed ParetoPilot framework to its core hyperparameters. Beyond reporting empirical observations, we provide mechanistic analyses grounded in the dynamics of diffusion models to explain how each parameter influences the generative trajectory.

\subsection{C.1 Rationale for Benchmark Selection and Metric Interpretation}

We conduct our fine-grained parameter sweeps on the \textbf{Synthetic Functions} and the \textbf{Real-world Engineering (RE) Suite}. This selection is strictly motivated by the statistical reliability of hyperparameter feedback across different tasks.

 Our statistical correlation tests reveal that performance variations on the Synthetic and RE suites exhibit highly significant linear consistency (Pearson $r = 0.70, p = 0.0027$ for Synthetic; $r = 0.61, p = 0.0121$ for RE). This demonstrates that the performance landscapes in these domains are smooth and predictable, allowing the marginal effects of hyperparameter variations to be systematically observed. Conversely, structured tasks like SciDesign exhibit massive empirical variation (Variance = 167.69, Range = 41.83\%) but lower statistical correlation with hyperparameter shifts (Pearson $p = 0.9248$). As discussed in the main text, this high variance stems primarily from the inherent extreme difficulty of generating fundamentally valid and meaningful solutions in these highly constrained search spaces. Consequently, the subtle performance variations induced by hyperparameter shifts in SciDesign and MO-NAS are easily overshadowed by the massive variance associated with structural validity. Therefore, performing sensitivity analysis on the Synthetic and RE suites provides a statistically sound and highly representative proxy for assessing the overall algorithmic performance trends.

 We evaluate the parameter sweeps using two primary metrics: Average Rank ($\downarrow$) and Percentage Improvement of HV ($\uparrow$). It is crucial to note that Average Rank evaluates relative standing among compared baselines. In problem settings where multiple algorithms perform similarly, microscopically small absolute performance differences can trigger disproportionately large rank fluctuations. \textbf{To counteract this relative sensitivity, we strongly encourage readers to focus on the Percentage Improvement of HV}. Because this metric is strictly anchored to the static offline dataset boundary ($HV(\mathcal{D})$), it provides a far more robust and absolute measure of the true generation quality and parameter sensitivity.

\begin{figure*}[htbp]
    \centering
    \begin{subfigure}[b]{0.48\textwidth}
        \centering
        \includegraphics[width=\textwidth]{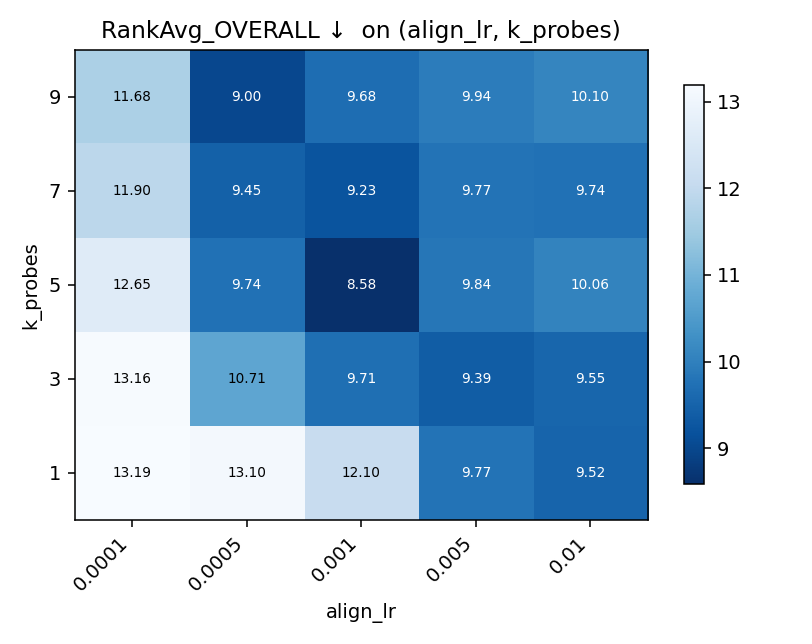} 
        \caption{Average Rank ($\downarrow$)}
    \end{subfigure}
    \hfill
    \begin{subfigure}[b]{0.48\textwidth}
        \centering
        \includegraphics[width=\textwidth]{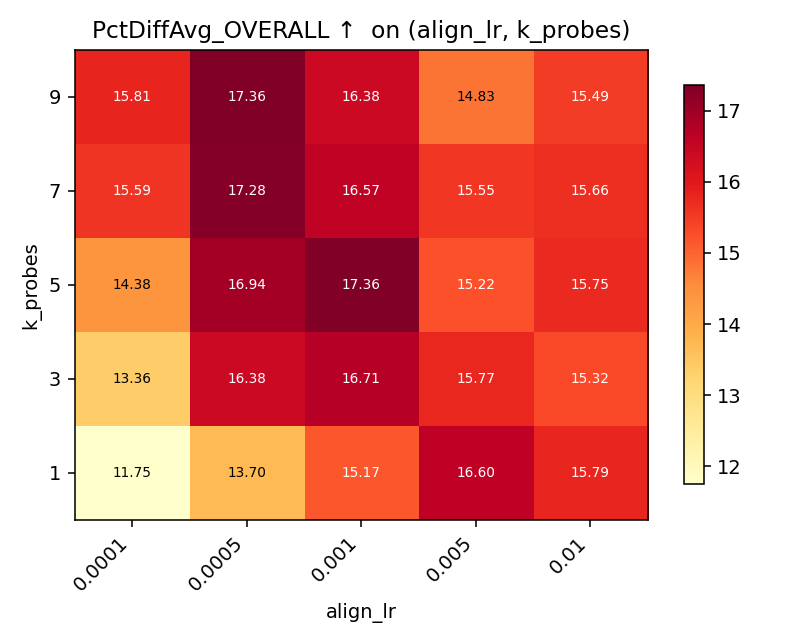} 
        \caption{Percentage Improvement of HV ($\uparrow$)}
    \end{subfigure}
    \caption{\textbf{Sensitivity of the INFER Phase.} Performance landscape over the alignment learning rate ($\eta$) and the number of probing steps ($k$).}
    \label{fig:sens_infer}
\end{figure*}

\subsection{C.2 The INFER Phase: Alignment Tracking ($\eta$ and $k$)}
\textbf{Parameter Role:} In the absence of external surrogates, ParetoPilot employs an un-reset Adam optimizer to infer the self-consistent objective anchor $y_t^*$. The alignment learning rate $\eta$ ('align\_lr') and the number of probing steps $k$ ('k\_probes') jointly determine the updating stride and effort expended at each inference interval to track the implicit data manifold.

\textbf{Empirical Observations:} As illustrated in Figure~\ref{fig:sens_infer}, reducing the tracking effort (e.g., $\eta = 0.0001$ or $k = 1$) results in a collapse in HV improvement, plunging to $\sim 11.75\%$. Conversely,  increasing the learning rate ($\eta \ge 0.01$) yields plateaued or slightly degraded HV ($\sim 15.49\%$), failing to reach the optimal peak. The system achieves maximum exploitation (HV $\sim 17.36\%$) at a moderate, balanced configuration of $\eta = 0.001$ and $k = 5$.

\textbf{Mechanistic Insight:} This trend reflects the challenge of tracking a moving target in a stochastic process. A low $\eta$ or small $k$ causes ''under-anchoring'': the optimizer's updates are too sluggish to keep up with the rapidly shifting valid data manifold across diffusion timesteps, leading to inaccurate directional bases. On the other hand, an excessively high $\eta$ induces overfitting to the instantaneous high-frequency Gaussian noise of a single timestep, causing the anchor to jitter wildly. The optimal configuration provides the precise amount of exponential moving average (EMA) momentum required to filter out step-wise noise while reliably tracking the underlying manifold.

\begin{figure*}[htbp]
    \centering
    \begin{subfigure}[b]{0.48\textwidth}
        \centering
        \includegraphics[width=\textwidth]{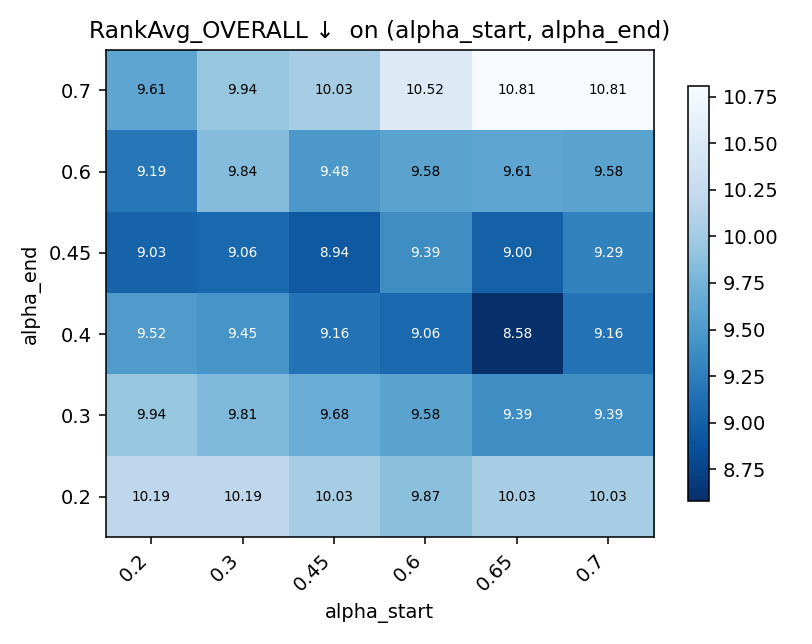}
        \caption{Average Rank ($\downarrow$)}
    \end{subfigure}
    \hfill
    \begin{subfigure}[b]{0.48\textwidth}
        \centering
        \includegraphics[width=\textwidth]{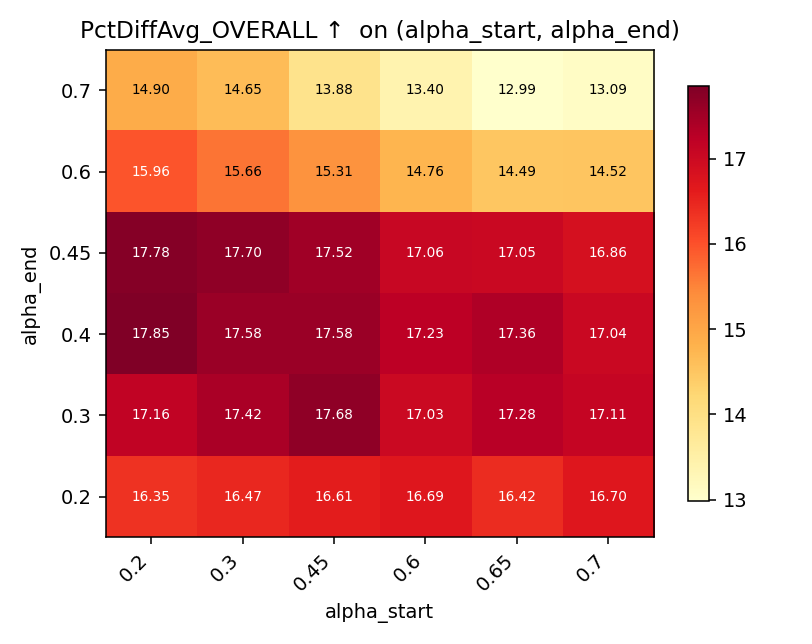}
        \caption{Percentage Improvement of HV ($\uparrow$)}
    \end{subfigure}
    \caption{\textbf{Sensitivity of the Dynamic Alpha Annealing.} Performance landscape over the perturbation blending weights $\alpha_{start}$ and $\alpha_{end}$.}
    \label{fig:sens_alpha}
\end{figure*}

\subsection{C.3 The PERTURB Phase: Dynamic Alpha Annealing ($\alpha_{start}$, $\alpha_{end}$)}
\textbf{Parameter Role:} The blending weight $\alpha_t$, annealed from $\alpha_{start}$ to $\alpha_{end}$, controls the critical trade-off between convergence (parallel gravity pushing toward the Pareto front) and diversity (orthogonal repulsion spreading the solutions).

\textbf{Empirical Observations:} Figure~\ref{fig:sens_alpha} reveals that maintaining a high $\alpha_{end}$ (e.g., $\alpha_{end} \ge 0.6$) is highly detrimental to HV, dropping performance to $\sim 13-14\%$. Decreasing the final weight to $\alpha_{end} \approx 0.40 \sim 0.45$ substantially improves HV to $> 17.5\%$. If the initial weight is suppressed ($\alpha_{start} \le 0.3$), the absolute HV increases (peaking at $\sim 17.85\%$ when $\alpha_{start}=0.2, \alpha_{end}=0.4$). However, this peak is accompanied by a severe deterioration in Average Rank ($> 9.5$). This divergence indicates that a low $\alpha_{start}$ yields exceptionally high performance on a narrow subset of tasks but severely sacrifices generalizability across diverse domains. The optimal robust configuration-securing the best Rank ($\sim 8.58$) while maintaining a highly competitive HV ($\sim 17.36\%$)—is achieved when $\alpha_{start}$ is moderately high ($0.45 \sim 0.65$) and $\alpha_{end}$ is low ($\sim 0.40$).

\textbf{Mechanistic Insight:} A low $\alpha_{start}$ implies weak early convergence gravity. While an early dominance of diversity (repulsion) might serendipitously scatter points to extreme high-HV regions on a few simpler tasks, it fails to provide the necessary macroscopic pull to escape sub-optimal local minima on complex tasks, leading to the observed rank degradation. The robust high-to-low $\alpha$ schedule guarantees broad generalizability: it strictly enforces macroscopic convergence when the diffusion noise is high, and safely pivots to lateral orthogonal expansion only as samples stably approach the true Pareto boundary.

\begin{figure*}[htbp]
    \centering
    \begin{subfigure}[b]{0.48\textwidth}
        \centering
        \includegraphics[width=\textwidth]{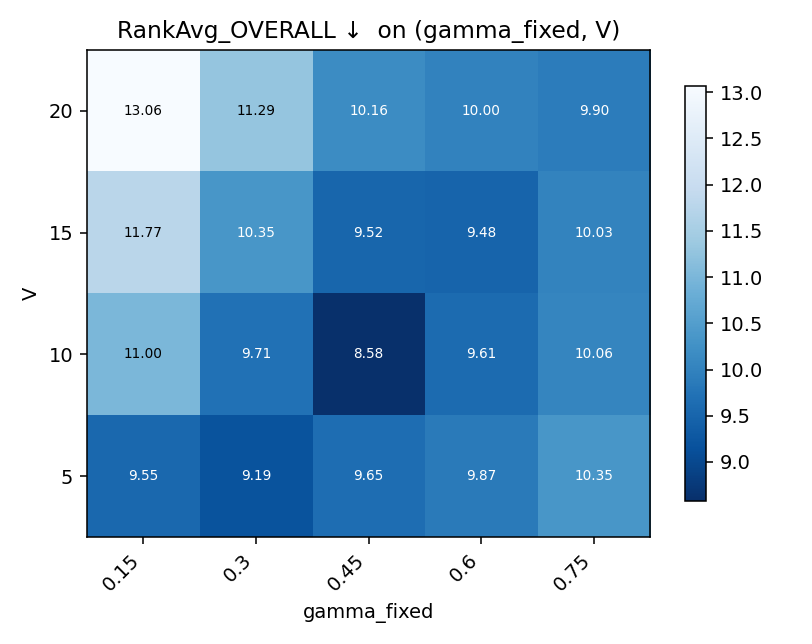}
        \caption{Average Rank ($\downarrow$)}
    \end{subfigure}
    \hfill
    \begin{subfigure}[b]{0.48\textwidth}
        \centering
        \includegraphics[width=\textwidth]{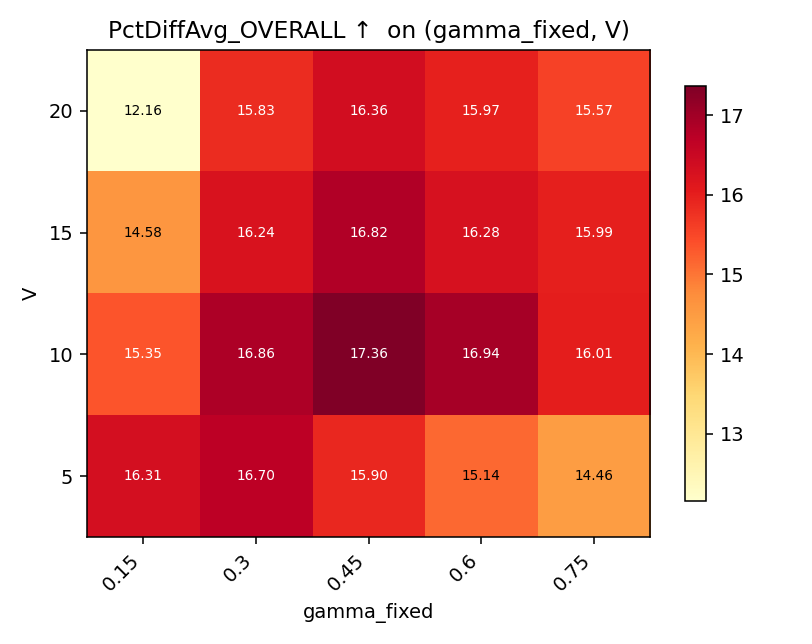}
        \caption{Percentage Improvement of HV ($\uparrow$)}
    \end{subfigure}
    \caption{\textbf{Sensitivity of IPG Frequency and Intensity.} Performance landscape over the IPG execution interval ($V$) and the extrapolation step size ($\gamma$).}
    \label{fig:sens_gamma}
\end{figure*}

\subsection{C.4 IPG Frequency and Extrapolation Intensity ($V$ and $\gamma$)}
\textbf{Parameter Role:} $V$ determines how often the kinematic perturbations intervene in the standard unconditional denoising process, while $\gamma$ governs the aggressive radius of the objective extrapolation. Together, they quantify the total guidance momentum applied to the generative model.

\textbf{Empirical Observations:} As shown in Figure~\ref{fig:sens_gamma}, executing IPG too frequently combined with large steps (e.g., $V = 5, \gamma \ge 0.6$) causes a noticeable deterioration in HV ($\sim 14.4\%$). Conversely, overly sparse and weak guidance ($V = 20, \gamma = 0.15$) fails to explore, yielding the lowest HV ($\sim 12.16\%$). The framework exhibits a distinct, robust diagonal sweet spot, centered at $V = 10$ and $\gamma = 0.45$, where the HV peaks at $17.36\%$.

\textbf{Mechanistic Insight:} Unconditional diffusion steps act as an implicit regularizer, pulling samples back to the valid data manifold learned from the offline dataset. A small $V$ combined with a large $\gamma$ constantly overrides this restoring force, compelling the neural network to extrapolate into wildly out-of-distribution (OOD) latent regions, which induces structural collapse and degraded evaluations. However, large $V$ and small $\gamma$ fail to generate sufficient momentum to traverse beyond the boundaries of the pre-collected dataset. The optimal configuration ($10, 0.45$) mathematically represents the precise equilibrium point where extrapolation velocity matches the manifold's natural restoring capacity.

\begin{figure*}[htbp]
    \centering
    \begin{subfigure}[b]{0.48\textwidth}
        \centering
        \includegraphics[width=\textwidth]{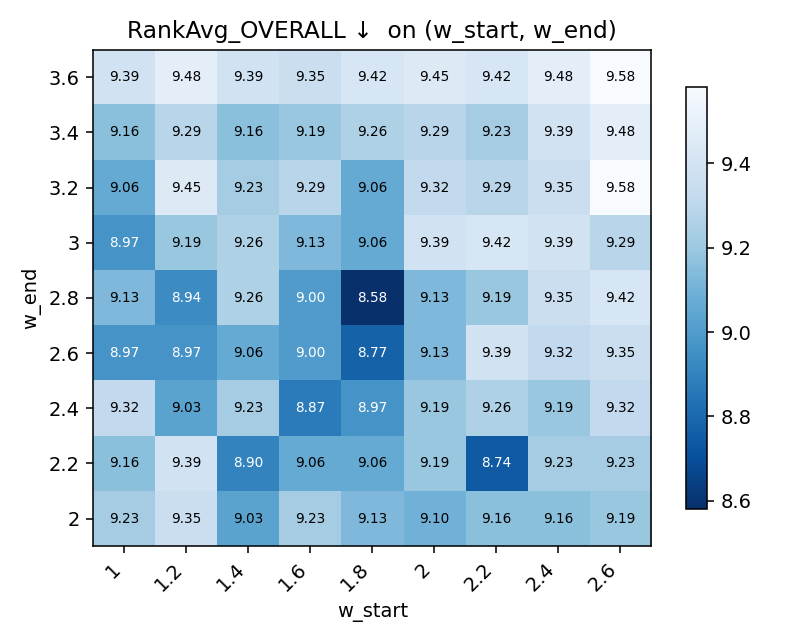}
        \caption{Average Rank ($\downarrow$)}
    \end{subfigure}
    \hfill
    \begin{subfigure}[b]{0.48\textwidth}
        \centering
        \includegraphics[width=\textwidth]{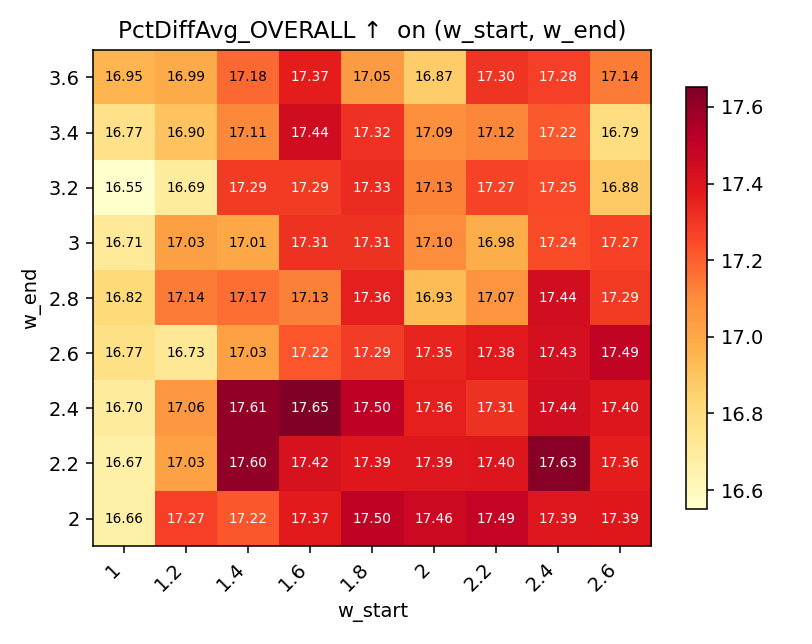}
        \caption{Percentage Improvement of HV ($\uparrow$)}
    \end{subfigure}
    \caption{\textbf{Sensitivity of CFG Extrapolation.} Performance landscape over the interpolated CFG weights $w_{start}$ and $w_{end}$.}
    \label{fig:sens_cfg}
\end{figure*}

\subsection{C.5 The GUIDE Phase: CFG Weight Interpolation ($w_{start}$, $w_{end}$)}
\textbf{Parameter Role:} The CFG scale translates the synthesized target condition into explicit modification of the neural noise prediction. The temporal interpolation from $w_{start}$ to $w_{end}$  regulates the generation guidance intensity over time.

\textbf{Empirical Observations:} As depicted in Figure~\ref{fig:sens_cfg}, the framework demonstrates a relatively robust and flat sensitivity landscape across the CFG parameters, maintaining competitive performance across most grid combinations. However, detailed spatial trends reveal key trade-offs: the Average Rank generally performs better in the bottom-left quadrant (moderate to low scales), whereas the absolute HV improvement peaks in the bottom-right quadrant (higher $w_{start}$, lower $w_{end}$). Overall, the system favors a relatively smaller final guidance scale $w_{end}$ (the lower half of the y-axis). Notably, when both initial and final guidance scales are excessively high (the top-right corner of the Rank heatmap), the Average Rank worsens noticeably ($> 9.3$). Conversely, when both guidance scales are too small (the bottom-left corner of the HV heatmap), the HV improvement hits its lowest point ($\sim 16.6\%$).

\textbf{Mechanistic Insight:} These two regimes clearly illustrate the fundamental trade-off between target extrapolation and manifold preservation. When the guidance intensity is too weak across the entire process (bottom-left), the model lacks the necessary kinetic momentum to escape the pre-collected offline dataset distribution, failing to push the trajectory toward the unobserved Pareto front, which explains the poor HV. On the other hand, when both $w_{start}$ and $w_{end}$ are extremely large (top-right), the overly aggressive guidance persistently overrides the unconditional prior (which is responsible for keeping the sample on the valid data manifold). This excessive force pushes the solutions completely off the structural manifold, leading to the generation of invalid or unreasonable designs, which severely degrades the overall ranking. Maintaining a moderate $w_{start}$ and a relatively small $w_{end}$ provides the perfect equilibrium: it generates sufficient momentum to discover novel Pareto optimal designs while safely landing the samples within the valid structural manifold.

\begin{figure}[htbp]
    \centering
    \includegraphics[width=\linewidth]{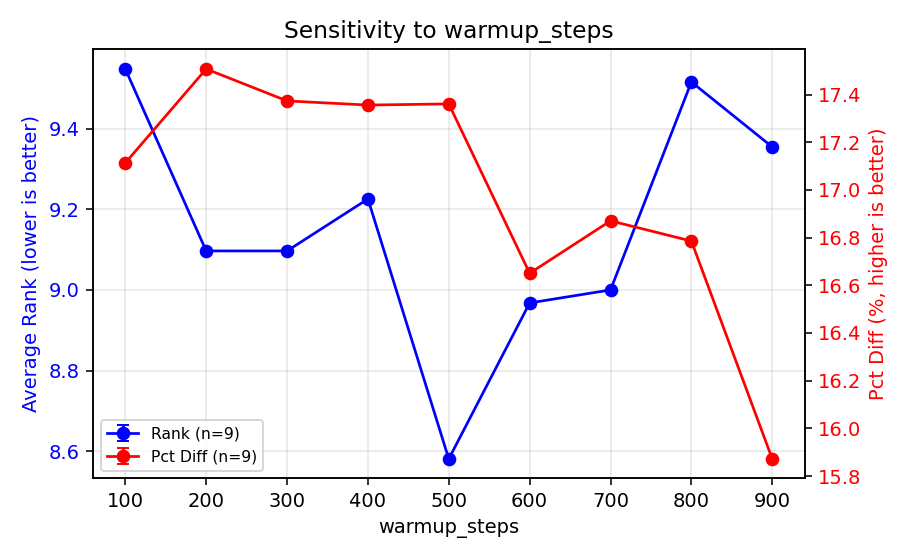} 
    \caption{\textbf{Sensitivity to Warmup Steps.} Impact of $T_{warmup}$ on Average Rank and HV Improvement.}
    \label{fig:sens_warmup}
\end{figure}

\subsection{C.6 Trajectory Stabilization: Warmup Steps ($T_{warmup}$)}
\textbf{Parameter Role:} $T_{warmup}$ defines the initial duration (in diffusion timesteps) during which the inferred anchors are purely tracked but not perturbed. It dictates when the active IPG steering commences.

\textbf{Empirical Observations:} The 1D line chart in Figure~\ref{fig:sens_warmup} directs our primary focus to the HV improvement (red line). Between $T_{warmup} \in [100, 500]$, the absolute HV improvement remains highly competitive and relatively stable (consistently $> 16.6\%$, peaking at $\sim 17.5\%$). However, extending the warmup beyond 500 steps causes a precipitous drop in HV (plummeting to $\sim 15.8\%$ at 900 steps). Critically, while the HV is resilient in the $100 \sim 500$ range, the Average Rank (blue line) deteriorates (worsening to $> 9.0$) for shorter warmups ($\le 400$). Thus, $T_{warmup} = 500$ emerges as the ideal sweet spot: it preserves the high HV ceiling while simultaneously ensuring optimal cross-task ranking stability ($\sim 8.58$).

\textbf{Mechanistic Insight:} The initial steps of reverse diffusion are characterized by an extremely low Signal-to-Noise Ratio (SNR). During this phase, the conditional alignment gradients are heavily obstructed by isotropic Gaussian noise. However, waiting too long ($T_{warmup} \ge 800$) leaves insufficient active denoising steps to meaningfully warp the latent manifold outward, resulting in the sharp drop in extrapolation HV. Allocating half the diffusion span ($T=500$) allows the un-reset Adam optimizer to fully filter the high-frequency initial noise via historical momentum accumulation, enabling highly precise orthogonal fluid navigation during the crucial mid-to-late structure-formation stages.

\section{Appendix D: Full Results of ParetoPilot}
\label{app:full_results}

This section reports the complete task-level hypervolume (HV) results on Off-MOO-Bench. While the main paper summarizes performance by domain-level average rank and percentage HV improvement, the following tables provide detailed results on individual benchmark tasks. Each entry is reported as mean (standard deviation) over repeated runs. The best result on each task is highlighted in \textbf{bold}, and the second-best result is \underline{underlined}.

These detailed results further support the main observations in Section~\ref{sec:experiments}. First, ParetoPilot does not rely on external objective regressors or preference classifiers, yet it remains competitive across both continuous synthetic functions and structured design tasks. Second, compared with methods that achieve strong results only on specific domains, ParetoPilot shows more consistent cross-domain behavior. Third, the results reveal the complementary failure modes of existing baselines: surrogate-based methods often perform well on smooth continuous tasks but degrade on discrete or structured domains, while preference- or reference-direction-based generative methods can be strong on selected tasks but exhibit substantial instability across domains.

Overall, the task-level results confirm that the advantage of ParetoPilot does not come from overfitting to a narrow subset of tasks. Instead, its inference-time IPG guidance provides a robust mechanism for steering a frozen conditional diffusion model toward high-quality Pareto regions while preserving the validity of generated designs.

\paragraph{Discussion.}
The full task-level results show that no single baseline dominates all benchmark families. Surrogate-based forward methods often obtain high HV on smooth continuous tasks, but their performance varies significantly on discrete and structured domains. Generative inverse methods also exhibit domain-specific behavior: ParetoFlow performs strongly on several RE tasks, PGD-MOO is competitive on selected structured tasks, and PCD achieves strong results on some synthetic and SciDesign tasks but suffers from large drops on several RE and C10MOP cases. In contrast, ParetoPilot consistently remains among the competitive methods across different benchmark families, which explains its strong overall rank in the main paper.

These results are consistent with our design motivation. By using unconditional denoising to keep samples close to the learned data manifold and applying IPG only as an inference-time condition perturbation, ParetoPilot avoids the deceptive extrapolation of external surrogates while still enabling active movement toward better Pareto regions.

\begin{table*}[htbp]
\caption{\textbf{Task-level HV results on DTLZ.} Each entry reports mean (standard deviation). The best result is highlighted in \textbf{bold}, and the second-best result is \underline{underlined}.}
\resizebox{\textwidth}{!}{
\begin{tabular}{lccccccc}
\toprule
 & DTLZ1-Exact-v0 & DTLZ2-Exact-v0 & DTLZ3-Exact-v0 & DTLZ4-Exact-v0 & DTLZ5-Exact-v0 & DTLZ6-Exact-v0 & DTLZ7-Exact-v0 \\
\midrule
HV-D(best) & 10.6026(0.0000) & 12.3101(0.0000) & 9.8969(0.0000) & 15.7150(0.0000) & 20.9752(0.0000) & 15.0487(0.0000) & 8.5649(0.0000) \\
PGD-MOO & 10.6444(0.0013) & 12.3894(0.0070) & 9.8882(0.0014) & 14.4686(0.0906) & 20.6654(0.1186) & 14.9128(0.2202) & 9.7574(0.1373) \\
End2End-GradNorm & 10.6454(0.0010) & 12.4422(0.0010) & 9.8865(0.0058) & 17.5746(0.0432) & 21.0523(0.0018) & 15.4163(0.0246) & 10.7046(0.0301) \\
End2End-PcGrad & 10.6437(0.0020) & 12.4442(0.0006) & 9.8957(0.0017) & 17.5728(0.0691) & \underline{21.0615(0.0007)} & 15.6090(0.0102) & 10.7211(0.0364) \\
End2End-Vanilla & 10.6446(0.0007) & 12.4434(0.0012) & 9.8311(0.0724) & \textbf{17.6593(0.0394)} & 21.0593(0.0017) & 15.6628(0.0159) & \underline{10.7783(0.0090)} \\
MultiHead-GradNorm & 10.6447(0.0007) & 12.4374(0.0030) & 9.8839(0.0031) & 17.5822(0.0285) & 20.8819(0.1212) & 15.5313(0.0424) & 10.7529(0.0347) \\
MultiHead-PcGrad & 10.6434(0.0004) & \textbf{12.4450(0.0003)} & 9.8893(0.0019) & 17.5383(0.0918) & 21.0579(0.0020) & 15.5867(0.0087) & 10.7469(0.0063) \\
MultiHead-Vanilla & 10.6447(0.0007) & 12.4444(0.0009) & 9.8888(0.0037) & 17.6079(0.1228) & 21.0599(0.0014) & \underline{15.6631(0.0076)} & 10.7753(0.0102) \\
MultipleModels-COM & 10.6331(0.0045) & 12.4121(0.0037) & \underline{9.8958(0.0008)} & 17.4699(0.0104) & 20.9802(0.0022) & 15.3323(0.0331) & 9.2330(0.0823) \\
MultipleModels-ICT & 10.6437(0.0017) & 12.4362(0.0014) & 9.8927(0.0022) & 17.3815(0.1126) & 20.9879(0.0143) & 15.4960(0.0125) & 10.7464(0.0076) \\
MultipleModels-IOM & 10.6434(0.0028) & 12.4427(0.0008) & \textbf{9.8959(0.0007)} & 16.7554(0.4177) & 21.0596(0.0008) & 15.5123(0.0027) & 10.4557(0.0727) \\
MultipleModels-RoMA & 10.6427(0.0012) & 12.4291(0.0043) & 9.8708(0.0073) & 17.2396(0.0806) & 20.9923(0.0202) & 15.4289(0.0144) & 10.5009(0.0439) \\
MultipleModels-TriMentoring & 10.6405(0.0020) & 12.4405(0.0022) & 9.2124(0.1049) & 17.5189(0.0863) & 20.9599(0.0337) & 15.6053(0.0138) & 10.6756(0.0439) \\
MultipleModels-Vanilla & 10.6454(0.0019) & \underline{12.4446(0.0009)} & 9.8898(0.0049) & 17.5831(0.0229) & \textbf{21.0619(0.0008)} & 15.6386(0.0069) & \textbf{10.7803(0.0030)} \\
PCD-reweight\_ref\_dir & \underline{10.6462(0.0005)} & 12.4136(0.0043) & 9.8913(0.0022) & \underline{17.6436(0.0245)} & 20.9289(0.0159) & 15.5211(0.0466) & 9.3522(0.0240) \\
ParetoFlow-Vanilla & 10.6038(0.0224) & 12.2750(0.0610) & 9.7913(0.0561) & 16.2309(0.5670) & 20.8071(0.0745) & 15.4051(0.0309) & 8.9699(0.1590) \\
ParetoPilot-Vanilla & \textbf{10.6474(0.0003)} & 12.3875(0.0068) & 9.8957(0.0026) & 17.6256(0.0287) & 20.4940(0.1518) & \textbf{15.7074(0.0787)} & 10.5282(0.0830) \\
\bottomrule
\end{tabular}

}
\end{table*}

\begin{table*}[htbp]
\caption{\textbf{Task-level HV results on ZDT.} Each entry reports mean (standard deviation).}
\resizebox{\textwidth}{!}{
\begin{tabular}{lccccc}
\toprule
 & ZDT1-Exact-v0 & ZDT2-Exact-v0 & ZDT3-Exact-v0 & ZDT4-Exact-v0 & ZDT6-Exact-v0 \\
\midrule
HV-D(best) & 4.1685(0.0000) & 4.6771(0.0000) & 5.1474(0.0000) & 5.4581(0.0000) & 4.6079(0.0000) \\
PGD-MOO & 4.5032(0.0466) & 5.3564(0.0427) & 5.6338(0.0579) & 5.0435(0.0996) & \textbf{4.8065(0.0212)} \\
End2End-GradNorm & 4.1782(0.0521) & 5.4559(0.0854) & 4.9859(0.0472) & 5.2640(0.1024) & 3.8893(1.0203) \\
End2End-PcGrad & 4.8397(0.0014) & 5.6241(0.0119) & 5.6675(0.0407) & 4.0616(0.3053) & 4.6133(0.1150) \\
End2End-Vanilla & 4.8396(0.0044) & 5.6205(0.0129) & 5.6517(0.0447) & 4.9036(0.2507) & 4.7725(0.0057) \\
MultiHead-GradNorm & 4.8295(0.0212) & 5.6256(0.0954) & 5.6520(0.0349) & 3.7142(0.1418) & 3.5259(0.9748) \\
MultiHead-PcGrad & 4.8353(0.0033) & 5.5175(0.0667) & 5.7111(0.1054) & 4.1136(0.2726) & 3.7291(1.3576) \\
MultiHead-Vanilla & \textbf{4.8475(0.0021)} & \underline{5.6260(0.0049)} & \textbf{5.8037(0.0484)} & 5.2118(0.0862) & \underline{4.7751(0.0027)} \\
MultipleModels-COM & 4.2664(0.0279) & 4.8421(0.0345) & 5.3502(0.1493) & 3.9724(0.1847) & 4.7494(0.0075) \\
MultipleModels-ICT & \underline{4.8436(0.0042)} & \textbf{5.6305(0.0161)} & 5.6197(0.0759) & 3.4559(0.1491) & 4.1169(1.0966) \\
MultipleModels-IOM & 4.3523(0.0489) & 4.9892(0.0496) & 5.5503(0.0153) & \underline{5.3392(0.0863)} & 4.7458(0.0043) \\
MultipleModels-RoMA & 4.8424(0.0039) & 5.6117(0.0357) & 5.7249(0.0297) & 4.9943(0.4336) & 3.2052(1.0467) \\
MultipleModels-TriMentoring & 4.4849(0.0094) & 5.3986(0.0108) & 5.3510(0.0205) & \textbf{5.3525(0.0215)} & 2.3288(0.3860) \\
MultipleModels-Vanilla & 4.8401(0.0057) & 5.6174(0.0125) & \underline{5.7501(0.0878)} & 5.1188(0.1595) & 4.7702(0.0064) \\
PCD-reweight\_ref\_dir & 4.3983(0.0303) & 5.0137(0.0400) & 5.4713(0.0390) & 5.2386(0.0866) & 4.6595(0.1329) \\
ParetoFlow-Vanilla & 4.1793(0.0522) & 5.4668(0.3014) & 5.2977(0.0684) & 4.8210(0.0887) & 4.5150(0.0378) \\
ParetoPilot-Vanilla & 4.5319(0.1073) & 5.1409(0.1760) & 5.6680(0.1733) & 5.2416(0.0215) & 4.6173(0.1213) \\
\bottomrule
\end{tabular}

}
\end{table*}

\begin{table*}[htbp]
\caption{\textbf{Task-level HV results on RE Suite, Part I.} Each entry reports mean (standard deviation).}
\resizebox{\textwidth}{!}{
\begin{tabular}{lcccccccc}
\toprule
 & RE21-Exact-v0 & RE22-Exact-v0 & RE23-Exact-v0 & RE24-Exact-v0 & RE25-Exact-v0 & RE31-Exact-v0 & RE32-Exact-v0 & RE33-Exact-v0 \\
\midrule
HV-D(best) & 4.0998(0.0000) & 4.7764(0.0000) & 4.7507(0.0000) & 4.5953(0.0000) & 4.7942(0.0000) & 10.6007(0.0000) & 10.5572(0.0000) & 10.5635(0.0000) \\
PGD-MOO & 4.4238(0.0171) & 4.8399(0.0002) & 4.8392(0.0002) & 4.8347(0.0009) & 4.8387(0.0007) & 10.6251(0.0160) & 10.6458(0.0013) & 10.4901(0.0591) \\
End2End-GradNorm & 4.4260(0.0322) & 4.8399(0.0000) & 4.8066(0.0373) & 4.5892(0.2284) & 4.8372(0.0039) & 10.6413(0.0061) & 10.6209(0.0223) & 10.5224(0.0694) \\
End2End-PcGrad & 4.5977(0.0006) & 4.8399(0.0000) & 4.8397(0.0002) & 4.8124(0.0196) & 4.8382(0.0010) & 10.6481(0.0001) & 10.6420(0.0037) & 10.6129(0.0020) \\
End2End-Vanilla & 4.3412(0.2833) & 4.8398(0.0000) & 4.8391(0.0009) & 4.2890(0.0156) & 4.3524(0.0000) & 10.6480(0.0001) & 10.6355(0.0107) & 10.6178(0.0050) \\
MultiHead-GradNorm & 4.4541(0.1958) & \underline{4.8399(0.0001)} & 4.7827(0.0404) & 4.6543(0.3615) & 4.8399(0.0000) & 10.6481(0.0001) & 6.7891(1.1541) & 10.5882(0.0223) \\
MultiHead-PcGrad & \textbf{4.5983(0.0002)} & \textbf{4.8399(0.0000)} & 4.7304(0.0748) & 4.8096(0.0331) & 4.8399(0.0000) & 10.6475(0.0005) & 10.6338(0.0002) & 10.6166(0.0030) \\
MultiHead-Vanilla & \underline{4.5982(0.0009)} & 4.8399(0.0001) & \underline{4.8398(0.0000)} & 4.5844(0.2656) & 4.8389(0.0018) & \underline{10.6481(0.0001)} & 10.6335(0.0055) & 10.6195(0.0018) \\
MultipleModels-COM & 4.0668(0.0642) & 4.7771(0.0239) & 4.8291(0.0012) & 4.6703(0.0607) & 4.8184(0.0022) & 10.5955(0.0020) & 10.6422(0.0027) & 10.5550(0.0041) \\
MultipleModels-ICT & 4.5962(0.0007) & 4.8399(0.0000) & 4.8252(0.0205) & 4.8359(0.0001) & 4.8399(0.0000) & 10.6475(0.0010) & 10.6467(0.0004) & 10.6077(0.0139) \\
MultipleModels-IOM & 4.5963(0.0008) & 4.8398(0.0000) & 4.8254(0.0249) & 4.8269(0.0072) & 4.8250(0.0110) & 10.6481(0.0000) & \underline{10.6477(0.0003)} & 10.6177(0.0044) \\
MultipleModels-RoMA & 4.4702(0.0821) & 4.8398(0.0001) & 4.8193(0.0141) & 4.6792(0.0625) & 4.8260(0.0185) & 10.6477(0.0006) & 10.4927(0.0514) & 10.5733(0.0255) \\
MultipleModels-TriMentoring & 4.5974(0.0015) & 4.8397(0.0001) & \textbf{4.8399(0.0000)} & 4.8361(0.0000) & \underline{4.8399(0.0000)} & 10.6476(0.0006) & 10.6463(0.0021) & 10.6117(0.0081) \\
MultipleModels-Vanilla & 4.5919(0.0068) & 4.8398(0.0000) & 4.8398(0.0000) & 4.4285(0.1157) & 4.7001(0.1346) & 10.6479(0.0004) & 10.6382(0.0076) & \underline{10.6208(0.0016)} \\
PCD-reweight\_ref\_dir & 4.3378(0.0059) & 4.8360(0.0032) & 4.8358(0.0030) & \underline{4.8800(0.0000)} & 4.8395(0.0002) & 10.6339(0.0063) & 10.6449(0.0011) & 10.5134(0.0207) \\
ParetoFlow-Vanilla & 4.1646(0.0816) & 4.8348(0.0652) & 4.8150(0.0253) & \textbf{5.0608(0.1250)} & \textbf{4.9644(0.0757)} & \textbf{10.9031(0.6774)} & \textbf{11.8628(0.9129)} & 10.5451(0.1064) \\
ParetoPilot-Vanilla & 4.4850(0.0747) & 4.8399(0.0001) & 4.8394(0.0006) & 4.7940(0.0658) & 4.4739(0.2428) & 10.6372(0.0047) & 10.5811(0.1286) & \textbf{10.6244(0.0018)} \\
\bottomrule
\end{tabular}

}
\end{table*}

\begin{table*}[htbp]
\caption{\textbf{Task-level HV results on RE Suite, Part II.} Each entry reports mean (standard deviation).}
\resizebox{\textwidth}{!}{
\begin{tabular}{lccccccc}
\toprule
 & RE34-Exact-v0 & RE35-Exact-v0 & RE36-Exact-v0 & RE37-Exact-v0 & RE41-Exact-v0 & RE42-Exact-v0 & RE61-Exact-v0 \\
\midrule
HV-D(best) & 9.2990(0.0000) & 10.0837(0.0000) & 7.6064(0.0000) & 5.5733(0.0000) & 18.2673(0.0000) & 14.5165(0.0000) & 97.4917(0.0000) \\
PGD-MOO & 9.5780(0.0933) & 10.3681(0.0542) & 9.6055(0.0921) & 6.0834(0.1049) & 18.9301(0.3667) & 18.8254(0.8518) & 104.1061(0.3561) \\
End2End-GradNorm & 9.9573(0.0103) & 10.3788(0.0227) & 9.6796(0.2453) & 6.6361(0.0200) & 20.7413(0.0734) & 21.3312(0.1512) & 104.0099(1.2785) \\
End2End-PcGrad & 10.1093(0.0036) & 10.5786(0.0059) & 10.1782(0.2257) & 6.7402(0.0028) & 20.7357(0.0484) & 22.6304(0.0971) & 103.9893(0.3574) \\
End2End-Vanilla & 10.1116(0.0013) & 10.5374(0.0142) & 9.9703(0.0946) & \underline{6.7443(0.0028)} & \underline{20.7745(0.0481)} & 22.4319(0.1870) & 108.8838(0.1491) \\
MultiHead-GradNorm & 9.7849(0.1226) & 10.4447(0.1011) & 8.7576(0.6094) & 6.6807(0.0303) & \textbf{20.7954(0.0418)} & 22.0513(0.4463) & 107.2638(0.6437) \\
MultiHead-PcGrad & 10.1135(0.0023) & 10.5776(0.0037) & 9.9955(0.1783) & 6.7393(0.0039) & 20.7566(0.0195) & 22.5663(0.1007) & 107.8201(0.3587) \\
MultiHead-Vanilla & 10.1125(0.0026) & 10.5582(0.0266) & 10.0681(0.0240) & \textbf{6.7456(0.0043)} & 20.7525(0.0425) & 22.4487(0.0450) & 108.7852(0.1759) \\
MultipleModels-COM & 9.7208(0.0462) & 10.4394(0.0387) & 8.1372(0.2190) & 6.2599(0.1035) & 19.2746(0.0664) & 17.0063(0.3322) & 107.6186(0.1062) \\
MultipleModels-ICT & 10.0866(0.0308) & 10.5709(0.0009) & \underline{10.2747(0.0190)} & 6.7106(0.0103) & 20.6707(0.0431) & \underline{22.7260(0.0523)} & 108.5381(0.1181) \\
MultipleModels-IOM & 10.0940(0.0052) & 10.5814(0.0032) & 10.2402(0.0881) & 6.7206(0.0100) & 20.6629(0.0487) & 22.0017(0.2385) & 108.1826(0.0642) \\
MultipleModels-RoMA & 10.0047(0.0277) & 10.5770(0.0075) & 9.2979(1.4627) & 6.6942(0.0127) & 20.6715(0.0680) & 21.5846(0.2635) & 108.3549(0.4294) \\
MultipleModels-TriMentoring & 10.0960(0.0113) & 10.4498(0.0676) & \textbf{10.2776(0.0620)} & 6.7414(0.0076) & 20.7608(0.0258) & \textbf{22.7506(0.1037)} & 108.6606(0.1581) \\
MultipleModels-Vanilla & \underline{10.1147(0.0035)} & 10.5175(0.0278) & 9.8719(0.1285) & 6.7411(0.0073) & 20.7570(0.0412) & 22.4670(0.0257) & \underline{108.9255(0.0965)} \\
PCD-reweight\_ref\_dir & 9.6090(0.0070) & 10.3553(0.0045) & 8.1215(0.0412) & 5.8964(0.0169) & 17.9104(0.0261) & 15.3261(0.0340) & 99.3675(0.0558) \\
ParetoFlow-Vanilla & \textbf{11.1952(0.3829)} & \textbf{10.6884(0.2554)} & 8.5821(0.5057) & 6.2855(0.7286) & 19.0137(0.7524) & 19.7284(4.9638) & 106.6319(11.6605) \\
ParetoPilot-Vanilla & 9.7408(0.1838) & \underline{10.5823(0.0201)} & 9.4095(0.5877) & 5.9066(0.1232) & 17.4280(0.9522) & 20.5428(0.5593) & \textbf{108.9726(0.1362)} \\
\bottomrule
\end{tabular}

}
\end{table*}

\begin{table*}[htbp]
\caption{\textbf{Task-level HV results on SciDesign.} Each entry reports mean (standard deviation).}
\resizebox{\textwidth}{!}{
\begin{tabular}{lccc}
\toprule
 & Molecule-Exact-v0 & Regex-Exact-v0 & ZINC-Exact-v0 \\
\midrule
HV-D(best) & 2.7320(0.0000) & 3.9609(0.0000) & 4.5220(0.0000) \\
PGD-MOO & \underline{3.4638(0.5746)} & \underline{6.4660(0.2988)} & \underline{4.6933(0.1073)} \\
End2End-GradNorm & 2.7281(0.2022) & 5.6775(0.5616) & 4.5421(0.0605) \\
End2End-PcGrad & 2.1574(0.6632) & 6.4121(0.3266) & 4.4768(0.0577) \\
End2End-Vanilla & 2.5665(0.1565) & 3.0951(0.2041) & 4.3370(0.2320) \\
MultiHead-GradNorm & 2.6193(0.1287) & 4.6595(0.9808) & 3.9395(0.5366) \\
MultiHead-PcGrad & 2.6122(0.2221) & 6.3390(0.2706) & 3.7981(0.6744) \\
MultiHead-Vanilla & 0.0000(0.0000) & 5.0375(1.1189) & 3.7409(0.3378) \\
MultipleModels-COM & 2.9141(0.0000) & 6.1367(0.2733) & 4.5607(0.0587) \\
MultipleModels-ICT & 2.7983(0.0000) & 6.1661(0.4128) & 4.2907(0.0324) \\
MultipleModels-IOM & 2.9769(0.1584) & 6.2588(0.0732) & 4.4748(0.0908) \\
MultipleModels-RoMA & 2.8847(0.0000) & 6.2475(0.3313) & 4.5505(0.0539) \\
MultipleModels-TriMentoring & 3.0066(0.2437) & 6.0594(0.3643) & 4.3950(0.1106) \\
MultipleModels-Vanilla & 2.6264(0.1515) & 2.9930(0.0000) & 4.2827(0.1865) \\
PCD-reweight\_ref\_dir & \textbf{4.0172(1.0631)} & 5.2791(0.4839) & \textbf{4.7117(0.0282)} \\
ParetoFlow-Vanilla & 1.8636(0.3845) & 5.1365(0.3347) & 4.3989(0.1355) \\
ParetoPilot-Vanilla & 3.3907(0.9714) & \textbf{6.7755(0.2618)} & 3.9625(0.6530) \\
\bottomrule
\end{tabular}

}
\end{table*}

\begin{table*}[htbp]
\caption{\textbf{Task-level HV results on OmniTest and VLMOP.} Each entry reports mean (standard deviation).}
\resizebox{\textwidth}{!}{
\begin{tabular}{lcccc}
\toprule
 & OmniTest-Exact-v0 & VLMOP1-Exact-v0 & VLMOP2-Exact-v0 & VLMOP3-Exact-v0 \\
\midrule
HV-D(best) & 4.5284(0.0000) & 0.0764(0.0000) & 1.7800(0.0000) & 45.6530(0.0000) \\
PGD-MOO & 4.7217(0.0336) & 0.0000(0.0000) & 2.2898(0.2674) & 45.4477(0.2812) \\
End2End-GradNorm & 4.6892(0.0248) & 0.3143(0.0012) & 3.9642(0.0589) & 44.4310(1.7364) \\
End2End-PcGrad & 4.7785(0.0055) & \underline{0.3168(0.0000)} & 4.2226(0.0198) & 45.8965(0.0185) \\
End2End-Vanilla & \underline{4.7813(0.0044)} & 0.3168(0.0000) & 4.1873(0.0182) & \underline{45.9350(0.0023)} \\
MultiHead-GradNorm & 3.4221(0.1213) & 0.2681(0.0081) & 3.2587(0.0445) & 42.4642(6.4543) \\
MultiHead-PcGrad & 4.1909(0.4427) & 0.3007(0.0108) & 4.1120(0.0544) & 45.9324(0.0070) \\
MultiHead-Vanilla & 4.7801(0.0055) & 0.2867(0.0462) & 4.0791(0.0353) & \textbf{45.9369(0.0008)} \\
MultipleModels-COM & 4.7428(0.0263) & \textbf{0.3169(0.0000)} & 1.8491(0.0255) & 45.9293(0.0036) \\
MultipleModels-ICT & 4.7672(0.0032) & 0.3168(0.0000) & \textbf{4.2390(0.0055)} & 45.9333(0.0003) \\
MultipleModels-IOM & 4.7812(0.0021) & 0.3083(0.0085) & \underline{4.2306(0.0060)} & 45.9329(0.0006) \\
MultipleModels-RoMA & 4.0859(0.1968) & 0.3167(0.0000) & 3.0275(0.4638) & 45.9053(0.0137) \\
MultipleModels-TriMentoring & 4.7584(0.0330) & 0.3168(0.0000) & 4.2261(0.0167) & 45.8911(0.0538) \\
MultipleModels-Vanilla & 4.7764(0.0102) & 0.3168(0.0000) & 4.1376(0.0469) & 45.9078(0.0342) \\
PCD-reweight\_ref\_dir & 4.7390(0.0149) & 0.0887(0.0094) & 2.9345(0.0912) & 45.9134(0.0052) \\
ParetoFlow-Vanilla & \textbf{4.7824(0.0045)} & 0.3164(0.0005) & 4.1311(0.1298) & 45.9174(0.0093) \\
ParetoPilot-Vanilla & 4.7616(0.0032) & 0.3156(0.0004) & 3.2575(0.3911) & 45.6093(0.2126) \\
\bottomrule
\end{tabular}

}
\end{table*}

\begin{table*}[htbp]
\caption{\textbf{Task-level HV results on IN1KMOP.} Each entry reports mean (standard deviation).}
\resizebox{\textwidth}{!}{
\begin{tabular}{lccccccccc}
\toprule
 & IN1KMOP1-Exact-v0 & IN1KMOP2-Exact-v0 & IN1KMOP3-Exact-v0 & IN1KMOP4-Exact-v0 & IN1KMOP5-Exact-v0 & IN1KMOP6-Exact-v0 & IN1KMOP7-Exact-v0 & IN1KMOP8-Exact-v0 & IN1KMOP9-Exact-v0 \\
\midrule
HV-D(best) & 6.5056(0.0000) & 7.4432(0.0000) & 17.6215(0.0000) & 4.2285(0.0000) & 4.5676(0.0000) & 11.3607(0.0000) & 5.6603(0.0000) & 13.5361(0.0000) & 18.7231(0.0000) \\
PGD-MOO & \textbf{5.7764(0.0448)} & \textbf{6.4469(0.0160)} & \underline{14.7927(0.1526)} & \textbf{4.4890(0.0370)} & \textbf{4.7680(0.0307)} & \textbf{11.9178(0.1320)} & \underline{5.2090(0.0691)} & 10.7653(0.0188) & 13.7921(0.4299) \\
End2End-GradNorm & 5.5355(0.0526) & 6.2112(0.0976) & 14.0854(0.0356) & 3.5142(0.2502) & 3.7056(0.0614) & 7.8456(0.3238) & 5.0300(0.1946) & 8.8447(0.3729) & 10.4017(0.4830) \\
End2End-PcGrad & 5.5880(0.0376) & 6.1930(0.0810) & 14.2204(0.0486) & 4.1951(0.0586) & 4.3336(0.0469) & 10.8289(0.1977) & 4.8131(0.1519) & 8.9314(0.2090) & 10.7061(0.4267) \\
End2End-Vanilla & 5.5064(0.0861) & 6.0265(0.3181) & 14.2612(0.0911) & 4.2583(0.0721) & 4.4967(0.0692) & 11.3668(0.1626) & 4.8069(0.0739) & 8.8601(0.4237) & 11.0895(0.5328) \\
MultiHead-GradNorm & 5.4521(0.1526) & 6.1347(0.1284) & 14.0546(0.1079) & 3.3337(0.7420) & 4.1189(0.4802) & 8.7521(0.4901) & 4.5465(0.1517) & 9.3022(0.8954) & 11.1799(0.5639) \\
MultiHead-PcGrad & 5.5772(0.0764) & 6.2293(0.0292) & 14.2016(0.1803) & 4.1828(0.0879) & 4.2695(0.0215) & 10.8771(0.2224) & 4.9243(0.2172) & 8.7081(0.3407) & 10.7318(0.2711) \\
MultiHead-Vanilla & 5.6303(0.0378) & 6.0423(0.0878) & 14.1507(0.0350) & 4.1758(0.0173) & 4.3544(0.0646) & 10.4462(0.2590) & 4.8226(0.1292) & 9.1052(0.1134) & 10.2628(0.9079) \\
MultipleModels-COM & 5.3152(0.1761) & 5.9750(0.3114) & 14.7706(0.0883) & 4.3316(0.0263) & 4.5868(0.0289) & 11.1593(0.0916) & 5.1289(0.1250) & \textbf{11.3849(0.2054)} & \textbf{14.8829(0.1091)} \\
MultipleModels-ICT & \underline{5.6811(0.0406)} & 6.2074(0.1284) & 14.2742(0.1467) & 4.2331(0.0302) & 4.4050(0.0462) & 11.2208(0.1444) & 5.0322(0.0133) & 9.6483(0.1442) & 11.7700(0.5626) \\
MultipleModels-IOM & 5.6013(0.0239) & 5.9875(0.2313) & 14.6665(0.1391) & 4.1468(0.3745) & 4.4353(0.1484) & 10.5515(1.0494) & 5.1384(0.1077) & \underline{11.1474(0.0394)} & \underline{14.7289(0.1955)} \\
MultipleModels-RoMA & 5.6296(0.0277) & 6.1633(0.2283) & 14.2070(0.0350) & 3.9835(0.3834) & 3.7188(0.0571) & 7.8305(0.2998) & 4.7207(0.0715) & 8.9892(0.1767) & 10.3739(0.7512) \\
MultipleModels-TriMentoring & 5.4907(0.0000) & 6.1038(0.1626) & 14.1112(0.1455) & 4.3959(0.0182) & 4.4666(0.0352) & 10.9612(0.1055) & 4.9301(0.0998) & 9.6745(0.8159) & 11.0046(0.5151) \\
MultipleModels-Vanilla & 5.6149(0.0270) & 5.8518(0.3624) & 14.3423(0.1682) & 4.2010(0.0459) & 4.4053(0.0671) & 10.5216(0.0912) & 4.7273(0.2190) & 9.2034(0.3871) & 11.0926(0.3796) \\
PCD-reweight\_ref\_dir & 5.5763(0.0920) & \underline{6.3523(0.0349)} & \textbf{14.9704(0.0887)} & \underline{4.4146(0.0286)} & \underline{4.6995(0.0355)} & \underline{11.6914(0.0379)} & \textbf{5.2194(0.0543)} & 10.7494(0.0868) & 13.6421(0.2005) \\
ParetoFlow-Vanilla & 5.3437(0.0781) & 6.1430(0.2273) & 14.3003(0.1335) & 4.3488(0.0452) & 4.6179(0.0305) & 11.6440(0.1480) & 4.8308(0.1092) & 11.1304(0.1653) & 13.6512(0.5981) \\
ParetoPilot-Vanilla & 5.5881(0.1262) & 6.3452(0.0271) & 14.4306(0.1511) & 4.3616(0.0144) & 4.6796(0.0262) & 11.5414(0.4110) & 4.9703(0.1145) & 9.6561(0.3777) & 12.4120(0.5935) \\
\bottomrule
\end{tabular}

}
\end{table*}

\begin{table*}[htbp]
\caption{\textbf{Task-level HV results on C10MOP.} Each entry reports mean (standard deviation).}
\resizebox{\textwidth}{!}{
\begin{tabular}{lccccccccc}
\toprule
 & C10MOP1-Exact-v0 & C10MOP2-Exact-v0 & C10MOP3-Exact-v0 & C10MOP4-Exact-v0 & C10MOP5-Exact-v0 & C10MOP6-Exact-v0 & C10MOP7-Exact-v0 & C10MOP8-Exact-v0 & C10MOP9-Exact-v0 \\
\midrule
HV-D(best) & 1.5138(0.0000) & 1.3944(0.0000) & 9.2114(0.0000) & 18.6226(0.0000) & 40.8036(0.0000) & 103.7284(0.0000) & 377.3130(0.0000) & 5.5662(0.0000) & 14.7710(0.0000) \\
PGD-MOO & 1.4572(0.0187) & 1.3476(0.0221) & \textbf{11.1712(0.0266)} & \textbf{23.9397(0.1132)} & \textbf{49.8997(0.0798)} & \textbf{107.5085(0.7179)} & \textbf{497.8848(2.3650)} & 5.4441(0.0468) & 14.2480(0.2394) \\
End2End-GradNorm & 1.3997(0.0895) & 1.2856(0.0601) & 9.5342(0.1755) & 21.2282(0.3953) & 40.1800(4.0619) & 88.4504(4.6618) & 361.1017(25.1773) & 5.1856(0.1909) & 13.2096(0.2864) \\
End2End-PcGrad & 1.2974(0.1066) & 1.1915(0.1032) & 9.9915(0.4438) & 21.6131(0.5158) & 34.9645(0.9921) & 92.3726(5.7208) & 347.9263(47.8850) & 4.9674(0.1645) & 12.5488(0.2649) \\
End2End-Vanilla & 1.3677(0.1272) & 1.3350(0.0265) & 9.9483(0.1191) & 21.9796(0.3839) & 34.9577(0.0536) & 94.2316(5.4055) & 324.6168(48.2092) & 5.1211(0.0714) & 12.2383(0.8091) \\
MultiHead-GradNorm & 1.4337(0.0271) & 1.2753(0.0516) & 10.0041(0.2973) & 21.1320(0.7665) & 30.9262(2.9519) & 80.4835(8.4592) & 310.6434(24.1791) & 5.1883(0.1627) & 13.1436(0.9136) \\
MultiHead-PcGrad & 1.4324(0.0257) & 1.2730(0.0738) & 10.4313(0.1742) & 21.1782(0.3644) & 37.4507(1.7936) & 90.4485(4.8829) & 359.6377(36.1944) & 4.9500(0.0986) & 12.5294(0.6879) \\
MultiHead-Vanilla & 1.4252(0.0191) & 1.3160(0.0405) & 10.1575(0.1704) & 21.3040(0.4665) & 36.6420(3.2406) & 90.9904(4.9047) & 319.2997(44.4659) & 5.1265(0.0957) & 11.7039(0.6321) \\
MultipleModels-COM & \underline{1.4702(0.0000)} & 1.3335(0.0067) & \underline{11.0891(0.0211)} & \underline{23.2346(0.3508)} & \underline{49.5871(0.2401)} & \underline{106.6173(0.7138)} & 485.0692(4.4976) & 5.2401(0.0447) & 14.0848(0.1408) \\
MultipleModels-ICT & 1.2701(0.1051) & 1.3250(0.0483) & 9.7963(0.1849) & 20.7535(0.5907) & 38.8178(0.7068) & 95.5371(5.2078) & 400.6956(48.2909) & 5.0122(0.0483) & 12.5872(0.3517) \\
MultipleModels-IOM & 1.4015(0.1169) & 1.3581(0.0316) & 10.5940(0.1225) & 21.9883(0.5028) & 48.1846(1.1751) & 101.9367(1.3865) & 467.0791(17.1030) & 5.0056(0.1474) & 13.6585(0.5442) \\
MultipleModels-RoMA & 1.4014(0.0808) & 1.3432(0.0187) & 10.2329(0.1806) & 21.3958(0.4573) & 37.5204(3.3477) & 93.0300(5.7819) & 380.0771(31.4465) & 5.1300(0.1126) & 13.0225(0.6888) \\
MultipleModels-TriMentoring & 1.4179(0.0344) & 1.1927(0.1263) & 9.9980(0.0764) & 20.7968(0.4244) & 38.3414(1.4263) & 95.4228(4.0905) & 401.3582(23.3829) & 4.9783(0.0796) & 12.5359(0.2976) \\
MultipleModels-Vanilla & 1.3624(0.0095) & 1.3579(0.0075) & 10.1262(0.0561) & 21.4364(0.3970) & 34.5111(1.1527) & 95.7866(7.3300) & 352.8379(27.0619) & 5.1750(0.0521) & 12.2175(0.2551) \\
PCD-reweight\_ref\_dir & \textbf{1.4940(0.0051)} & \textbf{1.3864(0.0084)} & 10.7980(0.1452) & 22.3490(0.3633) & 47.7173(3.1011) & 104.6372(1.9084) & 0.0000(0.0000) & \textbf{5.5545(0.0235)} & \textbf{14.6726(0.1150)} \\
ParetoFlow-Vanilla & 1.4438(0.0140) & 1.3437(0.0088) & 10.6566(0.1345) & 21.7095(0.6263) & 48.5272(0.9705) & 103.6718(2.6263) & 456.9208(22.3639) & 5.2680(0.1110) & 14.2886(0.4297) \\
ParetoPilot-Vanilla & 1.4422(0.0190) & \underline{1.3740(0.0150)} & 10.6607(0.1238) & 22.2172(0.6935) & 49.4100(0.3617) & 105.7011(0.8485) & \underline{490.7180(7.3722)} & \underline{5.5366(0.0441)} & \underline{14.5653(0.1861)} \\
\bottomrule
\end{tabular}

}
\end{table*}

\end{document}